\documentclass[10pt,twocolumn,letterpaper]{article}

\usepackage{cvpr}
\usepackage{times}
\usepackage{epsfig}
\usepackage{graphicx}
\usepackage{amsmath}
\usepackage{amssymb}

\usepackage{subfigure}

\usepackage{xfrac}

\def\eg{\emph{e.g}.,~} 
\def\ie{\emph{i.e}.,~}

\usepackage[pagebackref=true,breaklinks=true,letterpaper=true,colorlinks,bookmarks=false]{hyperref}

\cvprfinalcopy % *** Uncomment this line for the final submission

\def\cvprPaperID{7167} % *** Enter the CVPR Paper ID here

\ifcvprfinal\fi
\begin{document}

%%%%%%%%% TITLE
\title{Segmenting the Future}

\author{Hsu-kuang Chiu, Ehsan Adeli, Juan Carlos Niebles\\
Stanford University\\
{\tt\small \{hkchiu,eadeli,jniebles\}@cs.stanford.edu}
% \and
}

\maketitle
%\thispagestyle{empty}

%%%%%%%%% ABSTRACT
\begin{abstract}
Predicting the future is an important aspect for decision-making in robotics or autonomous driving systems, which heavily rely upon visual scene understanding.
%Semantic segmentation, on the other hand, is currently one of the most complete forms of visual scene understanding, in which the objective is to identify the semantic label for each pixel.
While prior work attempts to predict future video pixels, anticipate activities or forecast future scene semantic segments from segmentation of the preceding frames, methods that predict future semantic segmentation solely from the previous frame RGB data in a single end-to-end trainable model do not exist. In this paper, we propose a temporal encoder-decoder network architecture that encodes  RGB frames from the past and decodes the future semantic segmentation. The network is coupled with a new knowledge distillation training framework specific for the forecasting task. Our method, only seeing preceding video frames, implicitly models the scene segments while simultaneously accounting for the object dynamics to infer the future scene semantic segments. Our results on Cityscapes and Apolloscape outperform the baseline and current state-of-the-art methods. Code is available at 
%{\small \url{https://blinded_for_review/}}. 
{\small \url{https://github.com/eddyhkchiu/segmenting_the_future/}}.
\end{abstract}

%%%%%%%%% BODY TEXT
\section{Introduction}

Prediction of dynamics in visual scenes is one of the crucial components of intelligent decision-making in robotics and autonomous driving applications \cite{zhang2014predicting,gupta2018social,gui2018few}. To this end, learning useful representations that enable reasoning about the future has recently been of great attention. Example applications are anticipating activities \cite{vondrick2016anticipating}, predicting visual context \cite{zeng2017visual}, tracking to the future \cite{alahi2014socially}, forecasting human dynamics \cite{chiu2019action}, tracking dynamics in scenes \cite{luc2017predicting,luc2018predicting}, and predicting instance segments \cite{luc2018predicting}.

In recent years, semantic and instance segmentation of videos \cite{long2015fully,dai2016r,lin2017feature,zhang2018context} have become the leading methods to transform the scene into its semantic components, such as street, tree, vehicles, pedestrians, and obstacles. These semantic entities provide a high-level interpretation of the scene and hence predicting them can be of great interest. We argue that prediction of pixels in the RGB space is an overly perplexing task, while predicting high-level scene properties is sufficient, can be more useful, and is easier to interpret for decision-making purposes. Towards this direction, previous work predicted future semantic segments given the segmentation of the preceding frames \cite{rochan2018future,luc2017predicting}, or more sparsely predicted future instance segmentation from previous frames \cite{luc2018predicting}. In contrast, we (1) do not require segmentation of previous frames and (2) provide a dense forecast for all regions in the frame.
%Whereas, the most useful setting for interpretation of the future is to output future scene semantic segmentation given only the previous RGB frames, even though this is a much harder task compared to the former tasks. 

\begin{figure}[t]
    \centering
    \includegraphics[width=\linewidth]{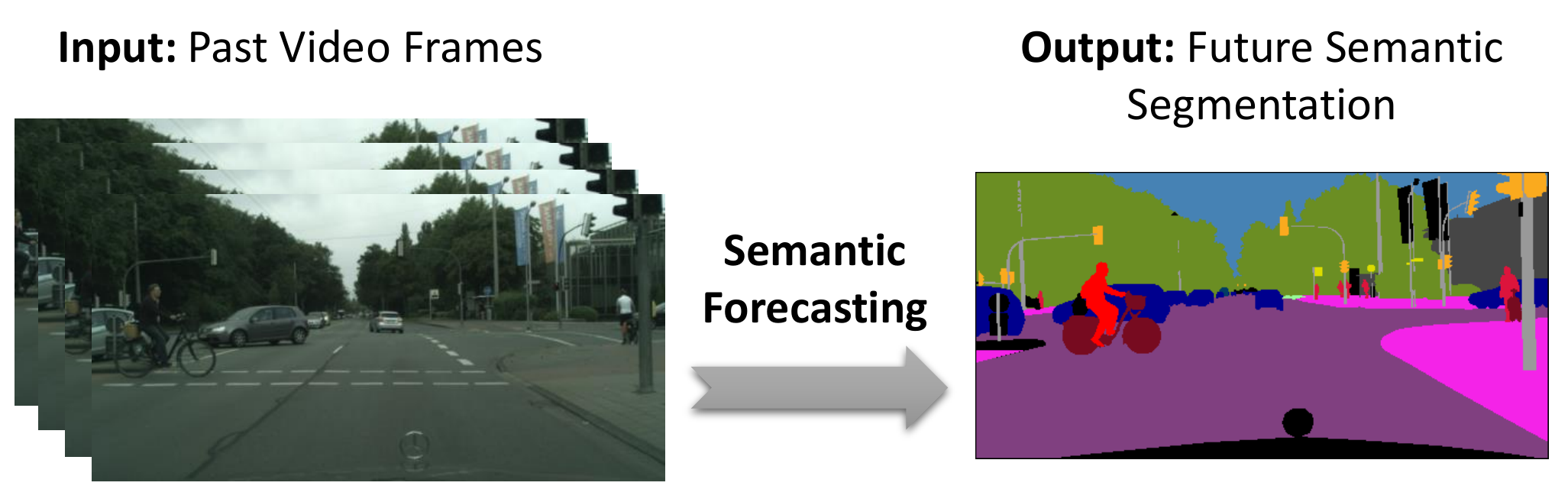} \vspace{-15pt}
    \caption{We obtain future semantic segmentation directly from preceding past frames in a single end-to-end trainable model. Our method implicitly infers the scene semantic segments while also forecasting the future configuration.}
    \label{fig:fig1}
    \vspace{-5pt}
\end{figure}

In this paper, we propose a model that predicts the future semantic segmentation in a video directly from pure RGB data of the previous frames (see Fig.~\ref{fig:fig1}). Inspired by the success of multi-resolution fully convolutional networks (FCN) \cite{long2015fully,dai2016r,lin2017feature,ronneberger2015u} % like U-Net \cite{ronneberger2015u}
segmentation, we propose an architecture that takes in a temporal sequence of frames and outputs the future semantic segmentation map. Unlike previous work \cite{zeng2017visual,kalchbrenner2016video,mathieu2016deep} that attempted to  predict future RGB video frames from preceding frames, our approach provides future semantic maps, which can enable reasoning about the forthcoming events. Alternatively, \cite{luc2017predicting} adopted a two-stage approach, first using the past RGB sequences to predict the future RGB frame, and then generating the future segmentation on top of that. %The performance of this approach is not satisfactory due to the difficulty of predicting future pixel values. 
On the contrary, one of our key observations is that future frame pixel values are \textit{not} necessary for generating future semantic segmentations, which is itself an easier task than generating future pixel values. We propose a single stage end-to-end trainable model that %does not require predicting the future RGB frame as an intermediate step. In this setting, our method 
learns to implicitly model the scene segments, and simultaneously account for the intrinsic dynamics of semantic maps for several object categories to predict future segmentation. In particular, this is a challenging task as objects in the semantic maps can significantly deform over the video frames due to changes in camera viewpoint, illumination, or orientation. To alleviate these challenges, our architecture encodes the sequence of input frames in a multi-resolution manner into a collective latent representation, and then decodes this representation gradually to the future semantic map. %Skip connections between the decoder and encoder at each resolution contribute to better recovery of the object boundaries.
%TODO: what gap needs fixing??%
We propose a novel knowledge distillation training framework that extracts future information to further refine the future semantic map. During the training stage, we utilize a fixed pre-trained single frame segmentation model and use it as a `teacher network.' Taking the future frame as the input, it  predicts the future segmentation. The predicted output from the teacher network provides additional information to guide the training of our main forecasting model, denoted by `student network.' This introduces one more training loss component, called distillation loss, which measures the difference between the outputs of the teacher and student networks. During inference, the teacher network is not used and the student network itself forecasts the future semantic segmentation using only the past RGB sequence as the input. %, without using any extra information. 
 With this new  distillation training framework, we can further improve the forecasting performance.

To evaluate the performance of our method, we use the Cityscapes \cite{cordts2016cityscapes} and the Apolloscape \cite{huang2018apolloscape} datasets %for predicting future video semantic segmentation 
under several scenarios, and compare our results with baseline methods. We predict the future semantic segmentation maps at three different temporal horizons (\ie, short-term, mid-term, and long-term) from the preceding RGB frames. Our method outperforms the previous methods although solving a much harder problem of predicting all semantic segments by only using past raw image sequences as input. Not only we define this harder problem and achieve the state-of-the-art, but we also outperform prior works under the simpler task that uses past semantic segmentation to forecast the future, even without modifying any part of our model.

In summary, our contributions are three-fold. First, we propose a single-stage end-to-end trainable model for the challenging task of predicting future semantic segmentation based on only the preceding RGB frames. 
%We propose a novel temporal fully convolutions encoder-decoder network with skip connections, tailored for the task of future segmentation in videos. 
Second, we propose a new knowledge distillation training framework that better uses future information. We introduce an additional distillation loss using a teacher network during training. %During inference, our main forecasting network still completes the same forecasting task by itself with the same type of input and output: forecasting future semantic segmentation purely based on past RGB sequence. This new training framework further improves the forecasting performance.
We show that our method can uncover the relations between the previous and future frames while taking the motion into account. %Our method does not need to predict the future frame's pixel values as an intermediate step. 
%Our proposed method does not require extremely costly semantic segmentation maps for the previous frames, in contrast, it predicts the future semantic map directly from preceding RGB frames.
Third, our proposed model also outperforms the previous state-of-the-art methods on the simpler setting that uses past semantic segmentation to forecast the future.

\section{Related Work}
%In this section, we discuss the literature most relevant to our work. First, we review the previous methods for segmentation in general with no elements of future prediction. Then, review recent works on forecasting and future prediction in the context of video processing applications.  

%\noindent\textbf{Semantic and Instance Segmentation.} 
\vspace{0pt}\noindent\textbf{Semantic and Instance Segmentation:} 
Semantic segmentation problems are often modeled by fully convolutional architectures (FCN) with multiple scales \cite{long2015fully,dai2016r,lin2017feature,ronneberger2015u,mathieu2016deep}, or by larger receptive fields \cite{zhou2018fine}. On the other hand, instance segmentation often maintains an strategy to generate instance-proposal regions \cite{ren2015faster} as part of the segmentation pipeline. Some early methods for this task integrate object detection and segmentation sequentially \cite{zagoruyko2016multipath,mosabbeb2007new}, and more recently in a joint multi-task end-to-end framework \cite{li2016fully}.

Other works have explored the utilization of temporal information and consistency across frames. Some of them are based on 3D data structures and operations \cite{zhang2010semantic} or CRF models \cite{kundu2016feature}, and others leveraged optical flow \cite{cheng2017segflow}. More recently, a number of methods utilize predictive feature learning techniques to enhance video segmentation. For instance, \cite{jin2017video} built a predictive model to learn features from the preceding input frames. All these works segment the last seen frame and do not evaluate the predictive power of the learned features for segmenting the future.

%\noindent\textbf{Video Forecasting.}
\vspace{0pt}\noindent\textbf{Video Forecasting:}
%There has been a growing interest in predicting and forecasting visual data. Such f
Visual forecasting tasks were defined as extrapolating video pixels to create realistic future frames  \cite{mathieu2016deep,zeng2017visual,bhattacharyya2018bayesian}. Although these methods have some success in predicting the future at the pixel-level, modeling raw RGB pixel values is rather cumbersome in comparison with predicting future high-level properties of the video. These high-level properties can not only be sufficient for analysis in many applications but also be more beneficial due to the higher level of semantic abstraction. Examples are anticipation of activities \cite{vondrick2016anticipating,kitani2012activity} or trajectories \cite{alahi2014socially}.

One of the best procedures for understanding and parsing scenes is scene semantic segmentation \cite{zhang2010semantic}. Recently, a few works proposed techniques for predicting semantic segmentation in videos. For instance, \cite{luc2017predicting} defined future semantic segmentation prediction through various configurations. They predicted future scene segmentation either from segmentation of the preceding frames or from the combination of segmentation and RGB data of the previous frames. They also presented a two-stage approach that first predicts the future frame pixel values, and then generates segmentation maps on top of the predicted future frame. Two other relevant works \cite{jin2017predicting, rochan2018future} predicted the future segmentation from previous frame segmentation maps. The former one \cite{jin2017predicting} developed a method based on flow anticipation using convolutional neural networks and the later one \cite{rochan2018future} developed a convolutional LSTM (ConvLSTM) model. Another work \cite{luc2018predicting}  developed a predictive model with fixed-sized features of the Mask R-CNN for future instance segmentation. Their work can only predict the future movement for limited types of objects, but not for other critical classes of importance for autonomous driving applications, such as roads and buildings. In summary, we mainly use X2X \cite{luc2017predicting} and ConvLSTM \cite{rochan2018future} as the baselines, since they are the closest works to ours. However, their models rely on sequences of past semantic segmentation as the input, or a two-stage approach by forecasting the future RGB frame as an intermediate step (see Table \ref{tab:related}). In contrast, we introduce an end-to-end trainable model for predicting the future segmentation solely based on the preceding RGB frames. %, which can be better suited for intelligent systems requiring to make instant decisions for the near future. 

\begin{table}[t!]
\small
\caption{Task setting comparison with prior work.
%: Luc \etal~\cite{luc2017predicting} and Nabavi \etal~\cite{rochan2018future}. %They focus on forecasting future semantic segmentation. 
We propose an end-to-end trainable model for forecasting future semantic segmentation (Seg) given only past RGB sequence. % as the input. %(*Two-stage approach, first predicting future RGB frame, then generating the segmentation.)
\vspace{-10pt}}
\label{tab:related}
\begin{center}
{\small 
\begin{tabular}{ lcc }
 \hline
 %milliseconds
  \textbf{Model} & \textbf{Input} & \textbf{Output} \\
 \hline \hline
  X2X \cite{luc2017predicting} & RGB  &  RGB \\
%  X2X* \cite{luc2017predicting} & RGB &  segmentation \\
  S2S \cite{luc2017predicting}& Seg  &  Seg \\
  XS2X \cite{luc2017predicting}& RGB+Seg  &  RGB \\
  XS2S \cite{luc2017predicting}& RGB+Seg  &  Seg \\
  XS2XS \cite{luc2017predicting}& RGB+Seg  &  RGB+Seg \\
 \hline
  ConvLSTM { \cite{rochan2018future}}&  Seg & Seg \\
 \hline
 Ours & RGB & Seg \\
 \hline
\end{tabular}
}
\end{center}
\vspace{-5pt}
\end{table}

\vspace{0pt}\noindent\textbf{Knowledge Distillation:} Knowledge distillation \cite{hinton2014distilling} was originally proposed to compress the knowledge from an ensemble of models into a single model during the training. This idea was extended to distill knowledge from different data modalities, such as optical flow and depth information for action recognition and video classification tasks \cite{luo2018graph,stroud2018d3d}. Different from the previous work, we propose a teacher network that takes the input from the same modality, the RGB frame, but in a different temporal range.

\section{Method}
\begin{figure*}[t!]
    \centering
    %\begin{tikzpicture}
    %\node[inner sep=0pt] (main) at (0,0)
    \includegraphics[width=1.0\linewidth]{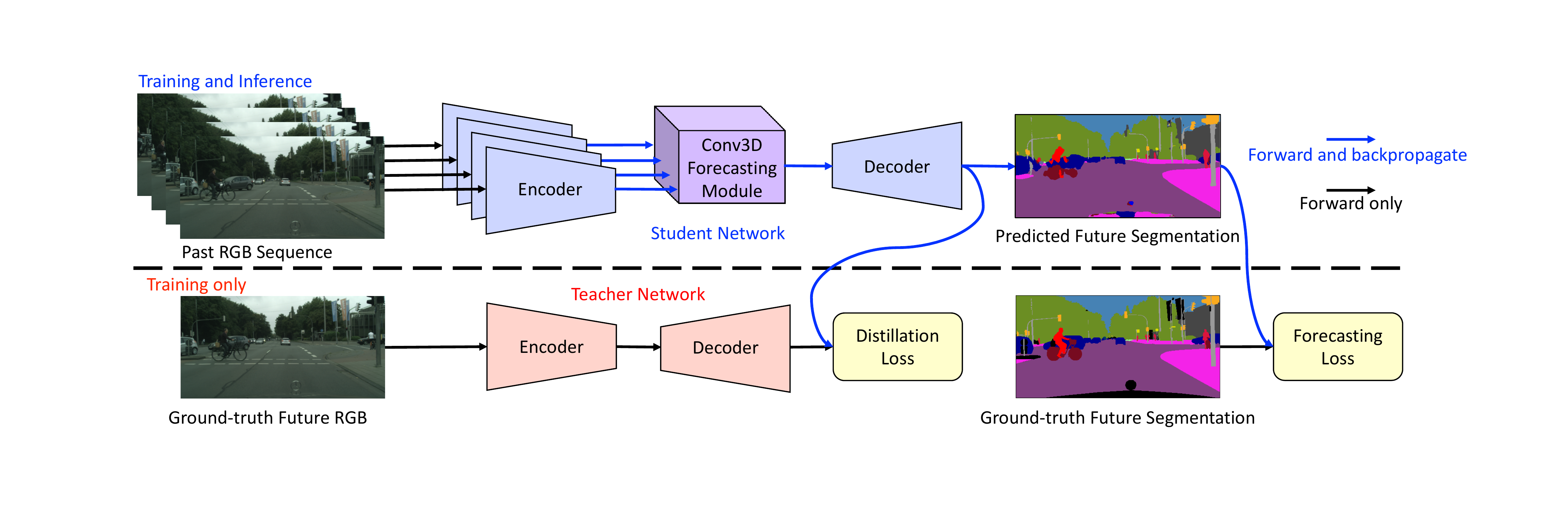} \vspace{-15pt}
    %{\includegraphics{images/arch_distill.pdf}};
    %\draw[] (-3.6,3.2) node[] {$X_{t-3d}$};
    %\draw[] (-2.5,3.2) node[] {$X_{t-2d}$};
    %\draw[] (-1.5,3.2) node[] {$X_{t-d}$};
    %\draw[] (-0.5,3.2) node[] {$X_{t}$};
    %\draw[] ( 3.5,3.2) node[] {$\hat{S}_{t+d'}$};
    %\end{tikzpicture}
    \caption{Architecture overview: our student network contains encoder, forecasting module, and decoder to forecast future semantic segmentation. During training, a fixed pre-trained teacher network takes the future RGB frame as input and predicts the future segmentation. The training loss is the weighted sum of the cross-entropy forecasting loss and the mean-squared error distillation loss. During inference, the teacher network is not used and the student network alone performs the forecasting task.}
    \label{fig:arch2}
\end{figure*}

Our goal is to build a model to forecast the future semantic segmentation given the past RGB sequence as the inputs. %To this end, we solve the semantic segmentation and forecasting problems jointly in a single end-to-end trainable model. 
Our proposed architecture involves two networks, the \textbf{student network} and the \textbf{teacher network}. The former performs our main forecasting task and the latter, during training, uses the future RGB frame to provide additional guidance to help the student network. At inference, only the student network is used to complete the semantic forecasting task, %, without knowing anything about the future RGB frame and the future segmentation 
 as shown in  Fig.~\ref{fig:arch2}.

The student network, as shown in the upper half of Fig.~\ref{fig:arch2}, has three main components: encoder, forecasting module, and decoder. The \textbf{encoder} generates feature maps in multiple resolutions from each input frame. For a video observed up to time $t$, the inputs are  $X_{t-3d}$, $X_{t-2d}$, $X_{t-d}$, and $X_{t}$, where $d$ denotes the displacement between each pair of the preceding frames. The \textbf{forecasting module} uses the feature maps (the lowest level maps from each past frame pathway) to learn a latent-space representation by consolidating temporal dynamics across them. This module uses a temporal 3D convolution structure and acts as a predictive feature learning module integrating feature maps from the preceding frames. Finally, the \textbf{decoder} combines the spatial feature maps (through skip connections to the encoder) and the temporal features (output of the Conv3D module) to generate the final semantic segmentation of the future frame at time $t+d'$. Note that $d'$ denotes the time delay in the future for which the semantic segmentation is sought. The ground-truth future segmentation is referred to by ${S}_{t+d'}$ and the prediction by $\hat{S}_{t+d'}$. The choice of $d'$ defines how far in the future we plan to segment. We experiment on three different settings of the combinations of $d$ and $d'$ for short-term, mid-term, and long-term semantic segmentation forecasting. 

The lower half of Fig.~\ref{fig:arch2} shows how the teacher network generates the additional loss term to help train the student network. Unlike X2X \cite{luc2017predicting} that used the future RGB frame as an intermediate training target, our model uses the future frame in a new knowledge distillation approach during training. The teacher network can be any fixed pre-trained single frame semantic segmentation network. It uses the future frame $X_{t+d'}$ as the input to predict the future semantic segmentation. The difference between the pre-softmax output features from the teacher network and the one from the student network is used as the additional training loss. During inference, %the teacher network and any information from the future are not used. The output of 
the student network alone is used to define the future segmentation as the output of our model. %More details of each of the student network's three modules and how the teacher network helps the student network during the training time will be explained in the following subsections.

%$X_{t-3d}$ $X_{t-2d}$ $X_{t-d}$ $S_{t+d}$ $\hat{S}_{t+d}$

\subsection{Student Network}
%This network is the main component of our model that performs the forecasting task during both training and inference time, as shown in the upper half of Fig.~\ref{fig:arch2}. 
This network contains three main components: past encoder, forecasting module, and future decoder to perform our main forecasting task. See the upper half of  Fig.~\ref{fig:arch2} and the supplementary material for more details. 

\noindent\textbf{Encoder:}
%Our forecasting architecture is inspired by the encoder-decoder FCN models  \cite{long2015fully,dai2016r,lin2017feature} 
%commonly used for segmentation tasks. %, such as U-Net \cite{ronneberger2015u}. 
In contrast to the previous semantic segmentation methods, which are based on encoder-decoder FCN models, our encoder module contains parallel pathways, one for each input preceding frame. 
Each pathway contains a series of fully convolutional networks, non-linearity layers, and max-pooling layers, to generate multi-resolution feature maps. 
The encoder can be designed using the common image classification models, such as VGG \cite{simonyan2014very} or ResNet \cite{he2016deep}, where the feature maps in different resolutions can be extracted right before each max-pooling layer. In our proposed method, we choose VGG19 with batch normalization as our encoder backbone. %For each input frame $X_i$, we first generate the feature map in the highest resolution by feeding it into two convolution modules, each of which contains a $3\times 3$ 2D convolutional layer (Conv2D), a batch normalization (BN) layer, and a rectified linear unit (ReLU). Then, the first feature map is fed into a max-pooling layer and another set of two convolution modules to generate the next feature map in a lower resolution. Similar operations are applied to generate all feature maps in different resolutions. 
Please refer to the supplementary material for the architecture details.

\noindent\textbf{Forecasting Module:}
%In contrast to most of the previous semantic segmentation models that only use two encoder and decoder modules, our model 
We introduce a forecasting module to learn predictive features and representations of temporal dynamics. For this purpose, we choose a 3D convolution network Conv3D to combine the encoded feature maps in the lowest resolution of each encoder pathway.
% (from each past time-step).
This is a simple design choice, but proves to be more accurate than LSTM \cite{gers1999learning} or ConvLSTM \cite{xingjian2015convolutional} in our experiments. The intuition behind using a Conv3D forecasting module is that it is able to learn more various motion dynamic information. On the contrary, the LSTM-based models always encode all the past information into a fixed size embedding at every timestep. The Conv3D module can access features from any past timesteps and learn to combine them in various ways to model the scene dynamics. This motivates our design of using Conv3D as the forecasting module. In addition to the 3D convolution pathway, skip connections provide another set of interactions between the encoder and the decoder.

\noindent\textbf{Decoder:}
The decoder takes the encoded feature maps at each resolution and the temporal dynamics representation as inputs and generates the future semantic segmentation, $\hat{S}_{t+d'}$. 
Different decoder structures are used in previous semantic segmentation models. 
% including FCN \cite{long2015fully} or Feature Pyramid Network \cite{lin2017feature} (FPN).
As an example, %the previous state-of-the-art semantic segmentation forecasting model, 
ConvLSTM \cite{rochan2018future}, reused the decoder of FCN \cite{long2015fully} with $1\times 1$ convolution, upsampling, and element-wise addition to obtain the segmentation at the original resolution.

We build our decoder symmetric to the encoder module sequence, as illustrated in the supplementary material. This choice gives us more computation capacity than the FCN decoder. Furthermore, the feature maps from different resolutions of the encoder can be directly concatenated with their counterparts in the decoder pathway %. Such skip connections have shown to be very useful for 
maintaining fine boundary information %as the features maps are gradually expanded in resolution 
\cite{ronneberger2015u,nie20183}. As a result of this symmetric structure, the lower level feature representation can be fed to the transpose convolution layer, which contains trainable parameters and upsamples the lower resolution feature to a higher resolution. Then, it is concatenated with the encoder feature maps of the corresponding resolution from the latest past time-step followed by 2D convolution modules. %Similar to the encoder, these sequence of modules are applied multiple times until we reach the original resolution.
% The detail architecture of the decoder can be seen in the supplementary material.

To construct the future segmentation from the decoder, the final convolution layer of the decoder generates a pre-softmax output tensor $O \in \mathbb{R}^{H \times W \times C}$, where $H$ and $W$ are the height and the width of the frames, and $C$ is the number of semantic categories. A soft-max on $O$ generates the predicted probability distribution tensor $P \in \mathbb{R}^{H \times W \times C}$:
{\small
\begin{equation}
P(h,w,c) = \frac {\exp(O(h,w,c))}{ \sum_{i=1}^C \exp(O(h,w,i)) },
\end{equation}
}where $h$, $w$ are the coordinates of a pixel, and $c$ is the index of each semantic class. %$O$ is the pre-softmax output tensor of the student network. $P(h,w,c)$ represents the predicted probability that the pixel at $(h,w)$ belongs to semantic category $c$. 
For each pixel $x = (h, w) \in \{H \times W \}$ and each semantic class $c$, we define the predicted probability function %$p_c(x)$:
% {\small
% \begin{equation}
    $p_c(x) = P(h, w, c)$,
% \end{equation}
% }
where $x = (h, w)$ represents the index of a pixel in the frame. To generate the final forecasting output, we choose the class with the highest probability as our prediction for that pixel. Hence, the predicted category for pixel $x = (h,w)$ at future time $t+d'$ is defined as:
% {\small
% \begin{equation}
$\hat{S}_{t+d'}(x)= \operatorname*{argmax}_c p_c(x),$
% \end{equation}
% }

\subsection{Training Loss and Teacher Network}
%Another important component of our model is the teacher network, as shown in the lower half of the Fig \ref{fig:arch2}. 
During the training, our loss $L$ is defined as the combination of the forecasting loss $L_f$ and the distillation loss $L_d$ using weighted sum $L = L_f + \lambda L_d$,
%{\small
%\begin{equation} \label{eq:loss}
%    L = L_f + \lambda L_d,
%\end{equation}
%}
where $\lambda$ is a hyperparameter. The forecasting loss $L_f$ measures the difference between the predicted output of the model $\hat{S}_{t+d'}$ and the ground-truth semantic segmentation $S_{t+d'}$. We use the cross-entropy function to define this classification loss:
{\small
\begin{equation}
L_f = -\frac{1}{HW}\sum_{x \in \{H\times W\}} \log \left(p_{g(x)}(x) \right),
\end{equation}
}where $g(x) \in \{1, \ldots, C\}$ defines the ground-truth label for pixel $x$. The second loss term $L_d$ is used to measure the difference between the outputs from the student network and the teacher network. %As mentioned earlier, 
The teacher network (lower half of Fig. \ref{fig:arch2}) is a fixed pre-trained single frame segmentation model. The teacher network has the same encoder and decoder architecture as the student network, but without sharing the trainable parameters. The teacher network takes the future RGB frame $X_{t+d'}$ as input and generates its own predicted future semantic segmentation $\hat{S}_{t+d'}^{te}$. Instead of directly comparing the predicted semantic segmentation, we define the distillation loss $L_d$ using the mean squared error between the pre-softmax output tensors from the student network and the teacher network:
{\small \begin{equation}  \label{eq:loss_distillation}
    L_d = \frac{1}{HWC}\sum_{h=1}^{H}\sum_{w=1}^{W}\sum_{c=1}^{C} (O(h,w,c) - O_{te}(h,w,c))^2,
\end{equation}
}
where $O$ is the pre-softmax output tensor from the student network, and $O_{te}$ is the one from the teacher network. With this distillation loss $L_d$, the single frame semantic segmentation teacher network provides additional regression guidance for the student network. During training, we minimize the overall loss $L$. During inference, the predicted output of the student network $\hat{S}_{t+d'}$ is the final output of our model. 

\begin{table*}[tb]
\small
\caption{Evaluation of our method in terms of mIOU, pixel-level accuracy (pAcc), and mean category accuracy (mAcc) in comparison with baseline and relevant methods on forecasting future semantic segmentation using past RGB sequence as the inputs. In each column, the best obtained results are typeset in boldface and the second best are underlined. %Our model outperforms the previous-state-of-the-art model ConvLSTM \cite{rochan2018future} and other baseline methods in all above evaluation metrics and all three forecasting time-ranges. 
\vspace{-1pt}}
\label{tab:res1}
\setlength{\tabcolsep}{5.25pt}
\renewcommand{\arraystretch}{0.7}
\begin{center}
\begin{tabular}{ lccccccccccc}
 \hline
 Model & \multicolumn{3}{c}{Short-term} & ~ & \multicolumn{3}{c}{Mid-term} & ~ & \multicolumn{3}{c}{Long-term} \\
 \cline{2-4} \cline{6-8} \cline{10-12}
% \hline
 & mIOU & pAcc & mAcc & ~ & mIOU & pAcc & mAcc & ~ & mIOU & pAcc & mAcc \\
 \hline
 Zero-motion & \underline{58.91} & \underline{91.96} & \underline{69.68} & ~ & \underline{48.15} & \underline{87.89} & \underline{59.67} & ~ & \underline{36.21} & \underline{81.77} & \underline{47.07} \\
 Two-stage & 49.17 & 90.22 & 61.68 & ~ & 26.53 & 74.60 & 36.19 & ~ & 9.64 & 44.86 & 14.49 \\
 X2X* \cite{luc2017predicting} & - & - & - & ~ & 23.00 & - & - & ~ & - & - & - \\
 ConvLSTM** \cite{rochan2018future} & 45.08 & 89.28 & 54.15 & ~ & 36.81 & 85.79 & 45.57 & ~ & 27.36 & 80.44 & 24.63 \\
  Ours & \textbf{65.08} & \textbf{93.83} & \textbf{74.36} & ~ & \textbf{56.98} & \textbf{91.38} & \textbf{67.67} & ~ & \textbf{40.81} & \textbf{86.03} & \textbf{50.13}\\
 \hline
 \multicolumn{12}{l}{*X2X is not the main focus of \cite{luc2017predicting}, and only the mid-term mIOU result is reported in this setting. } \\
  \multicolumn{12}{l}{**Our implementation of \cite{rochan2018future}, using past RGB sequence as the inputs.}
\end{tabular}
\end{center}
%\vspace{-7mm}
\end{table*}

\section{Experimental Results}
%In this section, w
%We evaluate our algorithm on a challenging and widely used dataset. We compare the results with several baseline and state-of-the-art algorithms. 

%\subsection{Dataset and Settings}
\noindent\textbf{Datasets:} We first evaluate our method on the \textit{Cityscapes} \cite{cordts2016cityscapes} dataset. %It is a large-scale dataset containing pixel-wise annotations for some frames. 
In the training, validation, and testing sets, the dataset provides $2975$, $500$, and $1525$ annotated frames with 19 semantic classes. 
%Another set of 20,000 coarsely annotated frames are also provided. 
In each of the video clips of length 30 (frames are indexed 0 to 29), the dataset provides fine annotations for the $19^\text{th}$ frame. In total there are 180,000 frames of resolution of $1024\times 2048$ pixels. Following the same setting of \cite{rochan2018future,luc2017predicting}, we only use the finely annotated frames and downsample the frames to resolution of $256\times 512$. We train our model using the Adam optimizer with initial learning rate of $0.001$, batch size 8,  and $\lambda=100$ (the two loss terms will have similar numerical scale in the beginning of training). %Besides, we initialize our teacher network by training a single frame segmentation model using the same dataset. 
We initialize the encoder of the student network with ImageNet \cite{deng2009imagenet} pre-trained weights.
%The previous works \cite{luc2017predicting,rochan2018future} only experimented on Cityscapes \cite{cordts2016cityscapes}. 

\textit{Apolloscape} dataset \cite{huang2018apolloscape} is also used as an additional experiment,
which has 140,000 frames with pixel-level annotations of 22 semantic classes. We extracted 1950 training and 380 validation sequences and down-sampled each frame to resolution of $320\times 384$. 
%We use the same time-range settings.
%, optimizer and other hyperparameters.

\noindent\textbf{Settings:}
%Our input is the past RGB frame sequence and we aim to forecast the future semantic segmentation. 
Following the forecasting settings in the related previous works \cite{luc2017predicting,rochan2018future}, we design three experimental settings on different time-ranges: short-term, mid-term, and long-term. For all settings, we always define the $19^\text{th}$ frame in the Cityscapes sequences, denoted by $S_{19}$, as the forecasting target frame, since the ground-truth semantic labels for this frame is available. The input RGB frames are selected from different timesteps in the past, denoted by $X_{i}$, depending on the forecasting time-range setting
%(\ie, short-term $d=d'=1$, mid-term $d=d'=3$, and long-term $d=3,d'=9$). 
For short-term forecasting, the input RGB frames are $X_{15}, X_{16}, X_{17}, X_{18}$; for mid-term forecasting $X_{7}, X_{10}, X_{13}, X_{16}$; and for long-term forecasting they are $X_{1}, X_{4}, X_{7}, X_{10}$, while the output for all is $\hat{S}_{19}$.

\noindent\textbf{Evaluation Metrics:}
Following previous work, we use the mean Intersection Over Union (mIOU) as the  performance metric for segmentation evaluation. We also report pixel-level accuracy (pAcc) and mean per-class accuracy (mAcc). Specifically, mIoU is the pixel IOU averaged across all classes; pAcc defines the percentage of correctly classified pixels; and mAcc is the average class accuracies.

\noindent\textbf{Baseline Methods:}
We use the ConvLSTM \cite{rochan2018future} model as the main comparison baseline, which outperforms the previous state-of-the-art S2S \cite{luc2017predicting}. This model achieves the state-of-the-art performance on a slightly different segmentation forecasting task, \ie, the inputs are the past segmentation sequences. %Although our experiment setting is different, which uses the past raw image frames as the inputs, ConvLSTM \cite{rochan2018future} can still be a strong baseline model. 
The architecture of ConvLSTM is based on the bidirectional ConvLSTM temporal module and uses the asymmetric Resnet101-FCN encoder-decoder backbone.  

Additionally, \cite{luc2017predicting} has an X2X architecture, which is also used as our baseline. It primarily focused on the same problem setting as the one of ConvLSTM \cite{rochan2018future}. But, they also presented an X2X \cite{luc2017predicting} architecture that uses the past RGB sequences to forecast the future RGB frame and then generate the future segmentation. We argue that it is not necessary to generate the future RGB frame as an intermediate step. Since \cite{luc2017predicting} only presents the result in the mid-term time-range setting, we implement a two-stage model based on the same idea. % of X2X \cite{luc2017predicting}. %, to complete the comparisons in all the three time-range settings. 
Another important baseline is denoted by `zero-motion.' This is the case that no motion is anticipated in the video and the future frame semantic segmentation is identical to that of the last observed frame. Although this is a very na\"ive baseline, it poses as a very challenging one \cite{martinez2017human,chiu2019action}, especially for short-term forecasting. To calculate this metric, we first train a single frame semantic segmentation model. Then, we apply it to the last input frame and use the segmentation map as the predicted result.% This baseline is neglected in the previous work. 

\subsection{Quantitative Results}
%\noindent\textbf{Semantic Segmentation Forecasting Results and Comparison:}
%\paragraph{Semantic Segmentation Forecasting Results and Comparison} 
Table \ref{tab:res1} shows the results of our method on the Cityscapes \cite{cordts2016cityscapes} dataset, compared with the previous state-of-the-art. % semantic segmentation forecasting model ConvLSTM \cite{rochan2018future}, the two-stage model X2X \cite{luc2017predicting}, and other baseline methods. 
Note that ConvLSTM \cite{rochan2018future} focused on predicting future segmentation directly from past semantic segmentation as the inputs. Therefore, we re-implemented their model, but training and testing using past RGB sequences as the inputs. As can be seen in Table \ref{tab:res1}, our model outperforms ConvLSTM  by a large margin in all the three time-ranges. That supports our design choice of using the symmetric encoder-decoder backbone and the Conv3D temporal module. % in our forecasting model. %We will discuss the reasons that our model performs better in more details in the ablation analysis section. Besides, our model also outperforms another baseline X2X \cite{luc2017predicting} that first forecasts the future RGB frame and then generates the future semantic segmentation. We further implement a two-stage model based on the same idea to complete the comparison in all the three time-range settings, and the results are shown in the second row of table \ref{tab:res1}. 
We can also see the significant performance difference between the two-stage models, including X2X \cite{luc2017predicting}, and our proposed single-stage model. The performance difference supports our argument that forecasting future RGB frame is not necessary for forecasting future semantic segmentation.  Besides, our method outperforms the the zero-motion baseline for all three time-ranges. Interestingly, the baselines perform worse than the zero-motion %na\"ive
baseline. 

\begin{figure}[t]
     \centering
     \includegraphics[width=\linewidth]{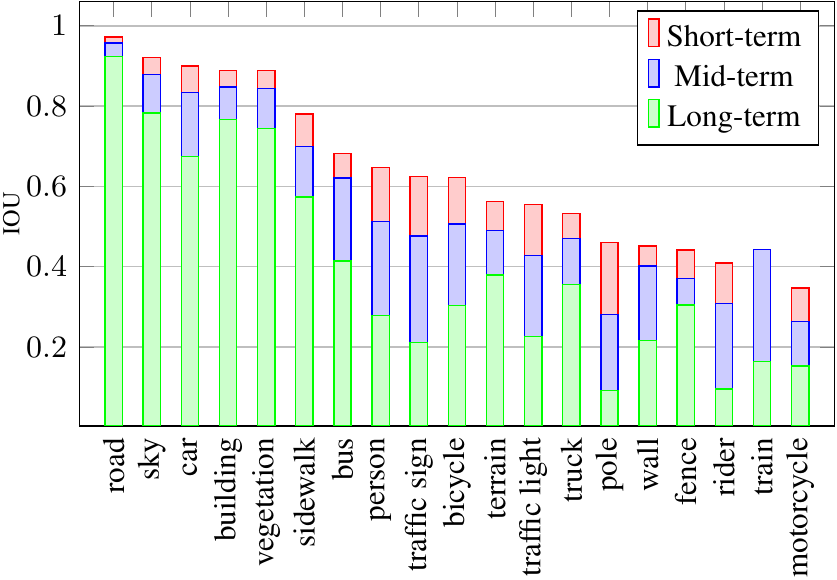}
     \caption{Per-class IOU for all 19 classes with respect to short, mid, and long-term forecasting. The forecasting performance varies a lot across different classes, implying that some classes are more difficult to correctly classify.
     }
     \label{fig:iou_per_class}
 \end{figure}

In addition to quantitative comparisons with previous works, we analyze our forecasting results for each of the 19 classes of objects. Fig.~\ref{fig:iou_per_class} shows the IOU comparison for all classes over three different forecasting time-ranges, sorted in descending orders. One can notice that the prediction accuracy varies a lot across different classes. %Some classes are relatively easier to forecast and segment than others.
To better understand the reasons why certain classes have lower IOUs, we calculate the confusion matrix for all these classes, as shown in Fig.~\ref{fig:conf_mat} (in the supplementary material). The x-axis in this figure refers to the predicted class labels and the y-axis represents the ground-truth class labels. %Each grid in the confusion matrix counts the number of pixels belongs to the associate ground-truth label, and predicted by our model to the associate prediction label. 
For instance, the `motorcycle' class, which had the lowest results in Fig.~\ref{fig:iou_per_class} is mainly confused with the `bicycle', `rider', and `car' classes. Additionally, we can see two light gray vertical lines for the `building' and `vegetation' predicted class labels. This shows that other classes are often mistaken with `building' and `vegetation'.

\begin{table*}[tb]
\small
\caption{Ablation results: evaluation of our method in terms of mIOU, pixel-level accuracy (pAcc), and mean category accuracy (mAcc) in comparison with variations of our method on forecasting future semantic segmentation using past RGB sequence as the inputs. In each column, the best obtained results are typeset in boldface and the second best are underlined.\vspace{-5pt}}
\label{tab:res2}
\setlength{\tabcolsep}{6pt}
\renewcommand{\arraystretch}{0.7}
\begin{center}
\begin{tabular}{ lccccccccccc}
 \hline
 Model &\multicolumn{3}{c}{Short-term} & ~ & \multicolumn{3}{c}{Mid-term} & ~ & \multicolumn{3}{c}{Long-term} \\
 \cline{2-4} \cline{6-8} \cline{10-12}
% \hline
 & mIOU & pAcc & mAcc & ~ & mIOU & pAcc & mAcc & ~ & mIOU & pAcc & mAcc \\
 \hline
 Ours w/o symmetric backbone  & 47.80 & 89.65 & 57.31 & ~ & 39.37 & 87.03 & 47.72 & ~ & 28.22 & 81.51 & 35.95  \\
 Ours w/ LSTM  & 48.22 & 90.26 & 58.39 & ~ & 39.57 & 86.93 & 49.18 & ~ & 26.61 & 81.46 & 33.19  \\
  Ours w/ ConvLSTM  & 58.62 & 92.92 & 68.85 & ~ & 47.96 & 89.29 & 58.24 & ~ & 34.44 & 84.21 & 43.09  \\
 Ours w/ Multi-Res Conv3D  & 59.24 & 93.24 & 69.44 & ~ & 49.33 & 90.38 & 59.39 & ~ & 36.96 & 85.42 & 45.98 \\
 Ours w/o distillation loss  & 63.60 & \underline{93.76} & 73.01 & ~ & 55.97 & 91.17 & 66.59 & ~ & 40.32 & 85.84 & 49.69\\
 %Ours w/ only lowest res. encoder distillation loss  &   &   &   & ~ &   &   &   & ~ &   &   &    \\
 %Ours w/ Multi-Res. encoder distillation loss &   &   &   & ~ & 55.18 & 91.00 & 65.87 & ~ &   &   &    \\
 %Ours w/ only encoder distillation loss & 63.27  & 93.59  & 72.69  & ~ & 55.18 & 91.00 & 65.87 & ~ & 40.07  & 85.76  & 49.61   \\
  Ours w/ 2 input frames & 63.85 & 93.67 & 73.30 & ~ & 55.67 & 91.15 & 66.58 & ~ & 40.37 & 85.75 & 50.02 \\
  Ours w/ 3 input frames & \underline{64.10} & 93.72 & \underline{74.18} & ~ & \underline{56.52} & \underline{91.27} & \underline{67.26} & ~ & \underline{40.77} & \underline{85.91} & \underline{50.11} \\
  Ours  & \textbf{65.08} & \textbf{93.83} & \textbf{74.36} & ~ & \textbf{56.98} & \textbf{91.38} & \textbf{67.67} & ~ & \textbf{40.81} & \textbf{86.03} & \textbf{50.13}\\
 \hline
% {\footnotesize *Our implementation} \\
 %\hline
 %Temporal-UNet (Ours) & RGB + Opticalflow & RomoveThisRow & 56.31 & RomoveThisRow \\
 %\hline 
\end{tabular}
% \begin{tablenotes}
%       \item *Our implementation
% \end{tablenotes}
\end{center}
\vspace{-2mm}
\end{table*}

% Workaround of above pgfplot memory issue
 
\noindent\textbf{Ablation:}
Table \ref{tab:res2} shows the ablation analysis results to examine  where the performance improvements derive from. The main differences between our model and previous work are the symmetric encoder-decoder backbone, the Conv3D temporal forecasting module, and the distillation training. 

First, to evaluate the symmetric encoder-decoder backbone design, we create another model by replacing our backbone architecture with the asymmetric one as in ConvLSTM \cite{rochan2018future}, which simply uses the decoder of FCN as the decoder (first row of table \ref{tab:res2}). We can see that all the evaluation metrics are significantly worse than our proposed model. %As mentioned earlier in the Method Section, o
Our proposed symmetric backbone decoder is designed with more computational capacity compared to FCN decoder. %That design decision is due to the fact that in our problem setting, t
The inputs and outputs represent two different types of information (image and segmentation). They are potentially far away from each other in the latent representation space. Therefore, more computation capacity is required, compared with previous works \cite{luc2017predicting,rochan2018future} whose inputs and outputs are all segmentation data.

%Now, we can conclude that the symmetric backbone design contributes significantly to  the performance gain. 
Next we analyze the impact of the temporal structures. We implement three other models by replacing our Conv3D temporal module with LSTM \cite{gers1999learning}, ConvLSTM \cite{xingjian2015convolutional} and the multi-resolution Conv3D. The results are reported in the second, third, and the fourth rows of Table \ref{tab:res2}. As suggested by the results, the LSTM module has significantly negative impact. ConvLSTM performs much better than LSTM, but still worse than our proposed Conv3D-based temporal module.
The multi-resolution Conv3D temporal module contains total five Conv3D layers, each of which takes the past feature maps from different resolution and generates the dynamic information for the decoder.
%Intuitively, we expect that this approach performs even better than our model.
%, which has only one temporal Conv3D layer processing the lowest resolution feature maps.
Empirically, we observe that its performance decreases. %(fourth row of the table)
One possible reason is that the larger number of trainable parameters makes the model prone to over-fitting. Furthermore, this approach requires more memory, which forces the model to operate on a smaller batch size. Therefore, the regularization effect of batch normalization becomes less effective.

Finally, we analyze the impact of the student-teacher architecture and the distillation training loss. Without the teacher network and the distillation training loss, results of our student-only model are shown in the fifth row of Table \ref{tab:res2}. %Compared with the above three ablation models, we can confirm that our symmetric encoder-decoder backbone and the Conv3D temporal module in the student network are essential to overall forecasting performance.
Our student-only model already outperforms other previous works shown in Table \ref{tab:res1}. 
%Remove encode distillation no space
% Furthermore, we implement a model by including a teacher network and a distillation loss calculated by the outputs of the encoders. The results become worse, as shown in the fifth row of the table. That implies that distilling knowledge only from the encoder output is not sufficient. On the contrary, 
%Our final proposed model uses the decoder output of the teacher network to define the distillation loss as in Equation \eqref{eq:loss_distillation}. 
Using the distillation loss of Eq. \eqref{eq:loss_distillation}, the mIOU scores further improve by 0.5\% to 1.5\% mIOU scores for all the three time-range settings%., compared with the student-only model.

Furthermore, we also experiment on using fewer numbers of input frames, and the results are shown in the sixth and seventh rows in Table \ref{tab:res2}. Using fewer numbers of input frames decreases the mIOU scores by 0.04\% to 1.31\%.

\vspace{0pt}\noindent\textbf{Forecasting Segmentation from Past Segmentation:}
In addition to our main experiments, we also experiment on another simpler task that forecasts future segmentation from past segmentation. This task is the exact problem setting that ConvLSTM \cite{rochan2018future} achieves the state-of-the-art performance. 
Our model still outperforms ConvLSTM\cite{rochan2018future} and another strong baseline, S2S\cite{luc2017predicting}, as in Table \ref{tab:res5}. 
% Too detailed, skip this part
%Notice that Cityscapes dataset does not provide ground-truth semantic segmentation masks for the past frames. ConvLSTM \cite{rochan2018future} uses PSPNet \cite{zhao2017pyramid} to generate their input segmentation sequences, and we use HRNet \cite{sun2019high} to generate our input. PSPNet provides more accurate semantic segmentation mask than HRNet (75.72 v.s. 74.28 in mIOU). However, our model still forecasts more accurate future semantics segmentation, which further shows that our model is more capable for the forecasting task.

\begin{table}[t!]
\caption{Comparison of mIOU with S2S \cite{luc2017predicting} and ConvLSTM \cite{rochan2018future} on forecasting segmentation using past segmentation as input.}
\label{tab:res5}
\setlength{\tabcolsep}{5.5pt}
\renewcommand{\arraystretch}{0.7}
\begin{center}
{%\small
\begin{tabular}{lccc}
 \hline
 Model & Short-term & Mid-term & Long-term \\
 \hline
 S2S & 62.60 & 59.40 & 47.80  \\
 ConvLSTM & 71.37 & 60.06 & -  \\
  Ours & \textbf{72.43}  & \textbf{65.53} & \textbf{50.52} \\
 \hline 
\end{tabular}
      }
\end{center}
\end{table}

\vspace{0pt}\noindent\textbf{Quantitative Results on Apolloscape Dataset:}
The previous works \cite{luc2017predicting,rochan2018future} only experimented on Cityscapes\cite{cordts2016cityscapes}. Additionally, we further experiment on Apolloscape\cite{huang2018apolloscape}, a
As shown in Table \ref{tab:res_apolloscape}, %with 1950 training sequences and 380 validation sequences. 
%Each image was down-sampled to resolution of $320\times 384$. We use the same time-range settings.
%, optimizer and other hyperparameters. 
our proposed model still outperforms ConvLSTM \cite{rochan2018future}. % with a smaller margin in Apolloscape. 
The distillation training further increases the forecasting performance. Notice that Apolloscape is more challenging compared with Cityscapes as discussed in \cite{huang2018apolloscape}, therefore we see smaller performance gains.

\begin{table}
\small
\caption{mIOU results on Apolloscape dataset.}
\label{tab:res_apolloscape}%\vspace{-10pt}
\setlength{\tabcolsep}{3pt}
\renewcommand{\arraystretch}{0.7}
\begin{center}
\begin{tabular}{lccc}
 \hline
 Model & Short-term & Mid-term & Long-term \\
 \hline
 Zero-motion  & 29.87 & 25.44 & 19.56 \\
 ConvLSTM**  & 28.27 & 23.29 & 17.30 \\
 Ours w/o distillation loss & 32.26 & 25.72 & 20.26 \\
 Ours & \textbf{32.58} & \textbf{26.09} & \textbf{20.47} \\
 \hline
  \multicolumn{4}{l}{**Our implementation of ConvLSTM with RGB inputs.} \\
\end{tabular}
\end{center}
\vspace{-15pt}
\end{table}

\subsection{Qualitative Results}
\begin{figure*}[tb]
\centering
  \subfigure[Short-term forecasting results]
  { \hspace{-8pt}\begin{tabular}{ccccc}
    \scalebox{0.75}{ $X_{t-3}, X_{t-2}, X_{t-1}, X_{t} $} & \scalebox{0.75}{ ${S}_{t+1}$: Ground-truth} &
    \scalebox{0.75}{ Ours} &
    \scalebox{0.75}{ ConvLSTM} & 
    \scalebox{0.75}{ Two-stage} \\
    
    \includegraphics[width=0.18\linewidth]{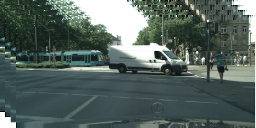} &
    \includegraphics[width=0.18\linewidth]{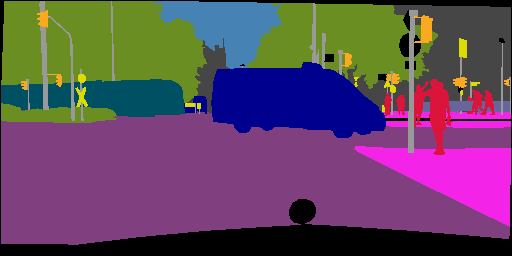} &
    \includegraphics[width=0.18\linewidth]{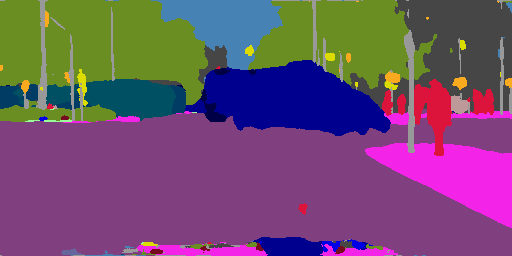} &
    \includegraphics[width=0.18\linewidth]{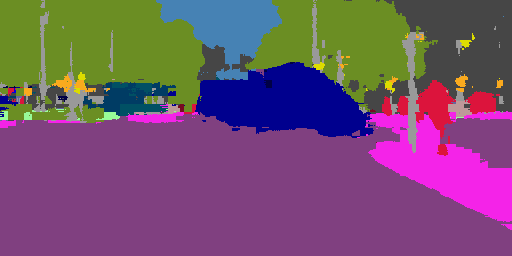} &
    \includegraphics[width=0.18\linewidth]{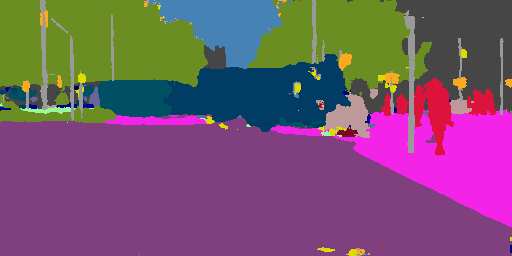}\\
    \includegraphics[width=0.18\linewidth]{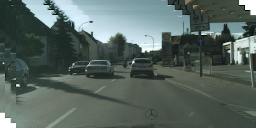} &
    \includegraphics[width=0.18\linewidth]{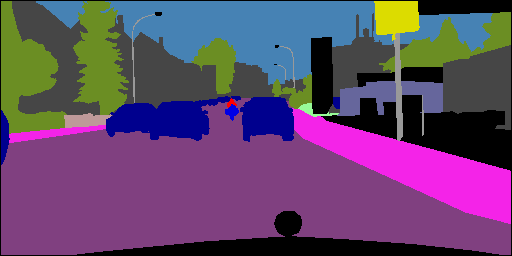} &
    \includegraphics[width=0.18\linewidth]{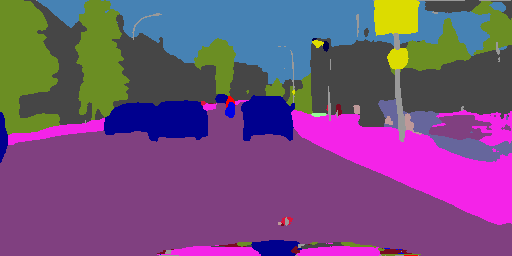} &
    \includegraphics[width=0.18\linewidth]{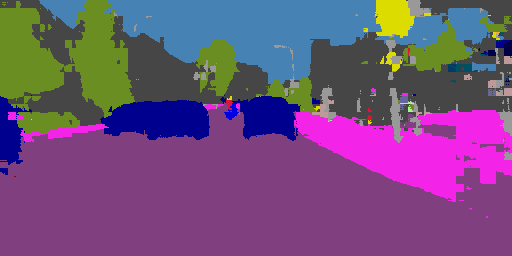} &
    \includegraphics[width=0.18\linewidth]{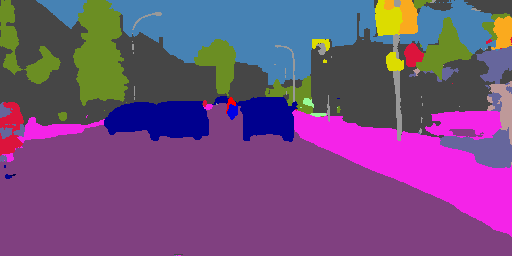}\\
    \end{tabular} \label{fig:short_qual}
  }\vspace{-0.5cm}
  \subfigure[Mid-term forecasting results]
  { \centering
  \renewcommand{\arraystretch}{0.7}
    \hspace{-8pt}\begin{tabular}{ccccc}
    \scalebox{0.75}{ $X_{t-9}, X_{t-6}, X_{t-3}, X_{t} $} & \scalebox{0.75}{${S}_{t+3}$ Ground-truth} & \scalebox{0.75}{ Ours} &
    \scalebox{0.75}{ ConvLSTM} & \scalebox{0.75}{ Two-stage} \\
   
    \includegraphics[width=0.18\linewidth]{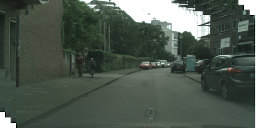} &
    \includegraphics[width=0.18\linewidth]{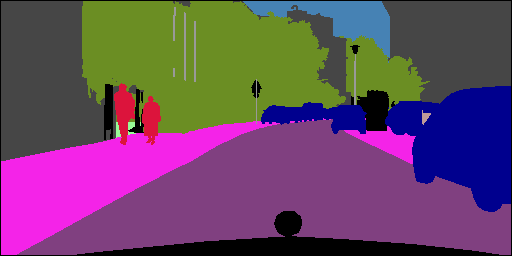} &
    \includegraphics[width=0.18\linewidth]{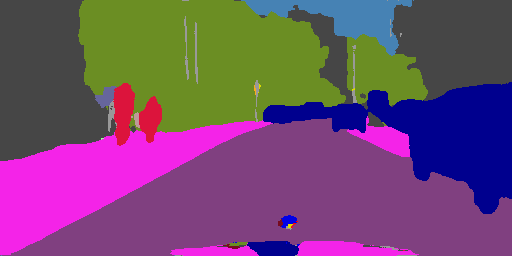} &
    \includegraphics[width=0.18\linewidth]{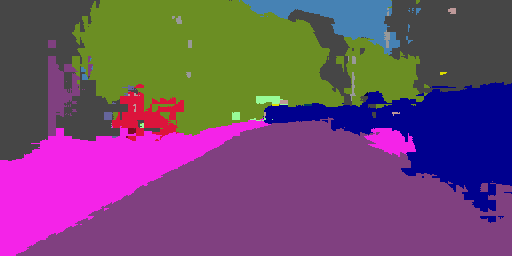} &
    \includegraphics[width=0.18\linewidth]{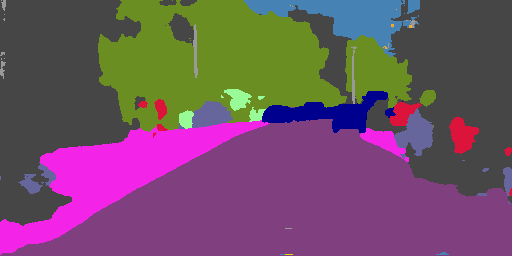}\\
    \includegraphics[width=0.18\linewidth]{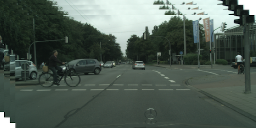} &
    \includegraphics[width=0.18\linewidth]{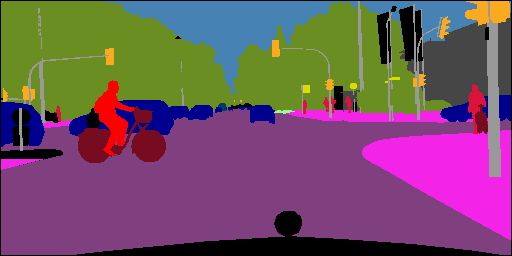} &
    \includegraphics[width=0.18\linewidth]{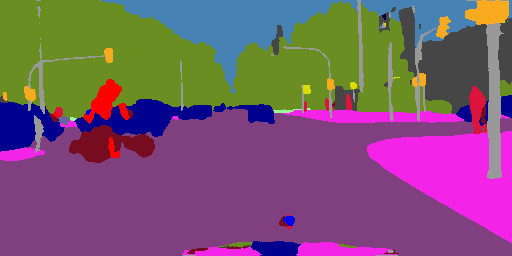} &
    \includegraphics[width=0.18\linewidth]{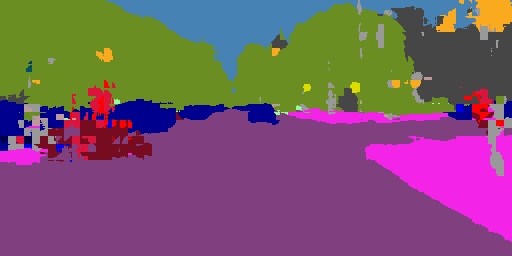} &
    \includegraphics[width=0.18\linewidth]{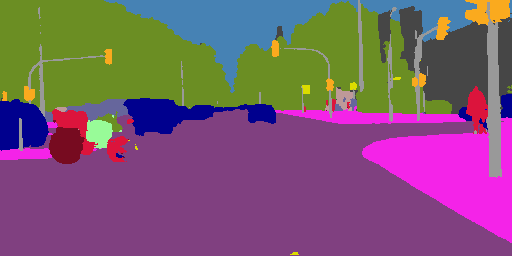}
    \end{tabular} \label{fig:mid_qual}
  }\vspace{-0.5cm}
  \subfigure[Long-term forecasting results]
  { \centering
    \hspace{-8pt}\begin{tabular}{ccccc}
    \scalebox{0.75}{ $X_{t-9}, X_{t-6}, X_{t-3}, X_{t} $} & \scalebox{0.75}{${S}_{t+9}$ Ground-truth} & \scalebox{0.75}{ Ours} &
    \scalebox{0.75}{ ConvLSTM} & \scalebox{0.75}{ Two-stage} \\
    \includegraphics[width=0.18\linewidth]{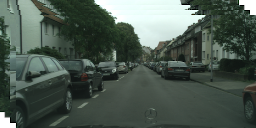} &
    \includegraphics[width=0.18\linewidth]{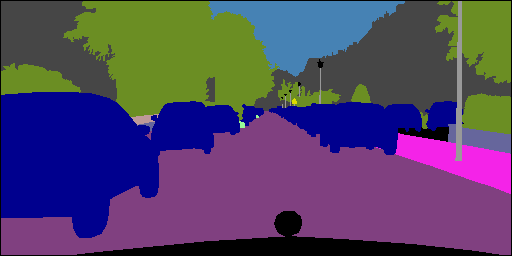} &
    \includegraphics[width=0.18\linewidth]{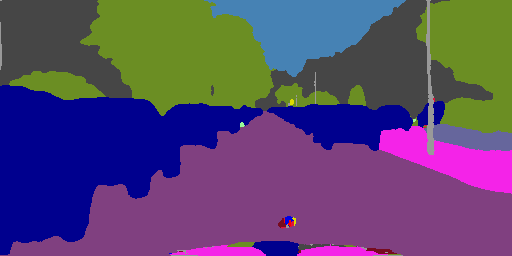} &
    \includegraphics[width=0.18\linewidth]{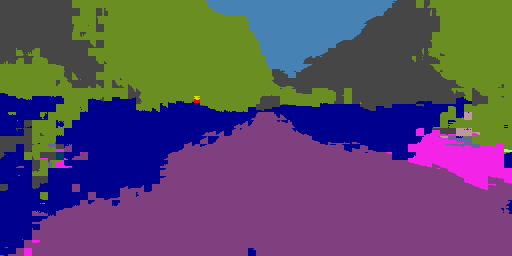} &
    \includegraphics[width=0.18\linewidth]{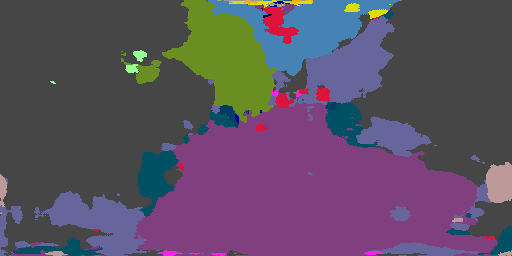}\\
    \includegraphics[width=0.18\linewidth]{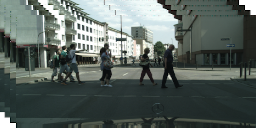} &
    \includegraphics[width=0.18\linewidth]{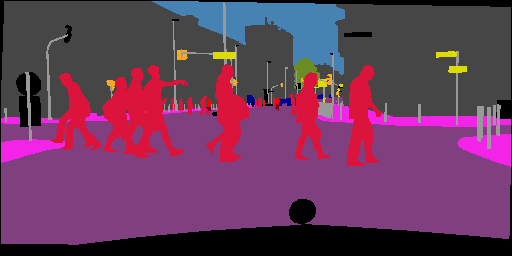} &
    \includegraphics[width=0.18\linewidth]{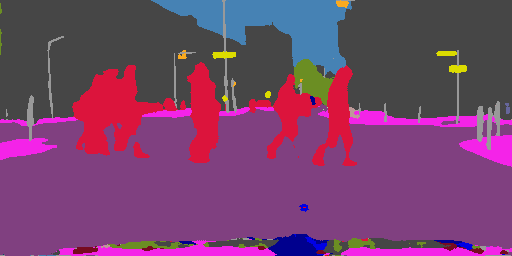} &
    \includegraphics[width=0.18\linewidth]{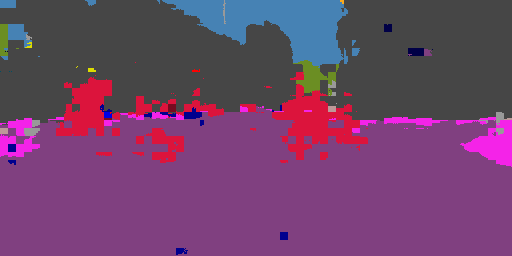} &
    \includegraphics[width=0.18\linewidth]{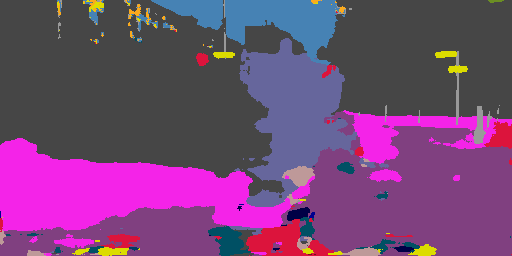}\\

    \end{tabular} \label{fig:long_qual}
  }
  \vspace{-10pt}
  \caption{Qualitative results. $\hat{S}$ denotes the predicted segmentation and $S$ the ground-truth. %Each row corresponds to a different video. %In all three cases, the semantic segmentation prediction is done using the preceding four frames with different lengths of time horizon in the future, \ie, $d'=1$, 3, and 9 for short, mid, and long-term forecasting, respectively. 
  See supplementary material for more results.}
  \label{fig:qual}
  \vspace{-0.3cm}
\end{figure*}

\noindent\textbf{Mid-term Forecasting:} We start this section with the mid-term forecasting results, as this is the most widely used setting in the previous work for evaluation.
Fig.~\ref{fig:mid_qual} shows the qualitative results, which uses the past RGB sequence $X_{7}, X_{10}, X_{13}, X_{16}$ to forecast the future semantic segmentation at time-step 19, denoted as $S_{19}$. In this figure, each row is a separate sample sequence, and the left most column is the past RGB input sequence, followed by the ground-truth future segmentation, our predicted future semantic segmentation
and two other baselines.
% in the third column, ConvLSTM \cite{rochan2018future} results in the fourth column, and the two-stage method results in the last column. 

\begin{figure*}[!h]
    \centering
    \begin{tabular}{cccc}
    \scalebox{0.75}{$X$} & \scalebox{0.75}{$\hat{S}_{t+1}$} & \scalebox{0.75}{$\hat{S}_{t+3}$} & \scalebox{0.75}{$\hat{S}_{t+9}$} \\
    \includegraphics[width=0.2\linewidth]{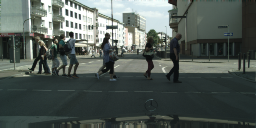} &
    \includegraphics[width=0.2\linewidth]{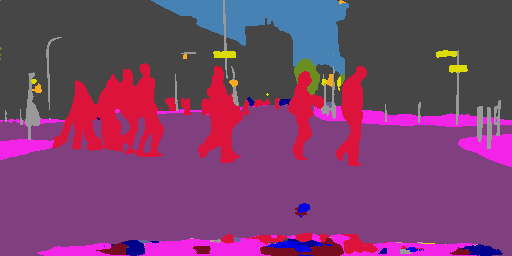} &
    \includegraphics[width=0.2\linewidth]{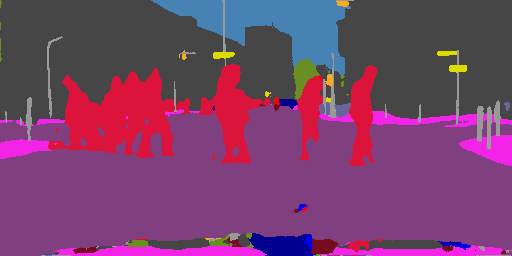} &
    \includegraphics[width=0.2\linewidth]{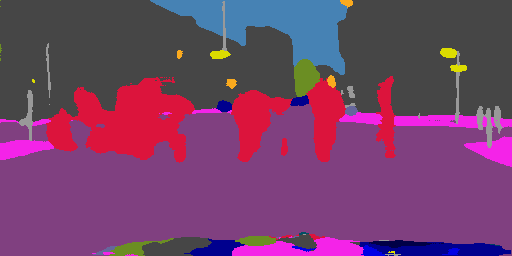}\\
    \end{tabular}\vspace{-0pt}
    \caption{Forecasting results of three different time-range settings for the same sample. $X$ is the last RGB frame; $\hat{S}_{t+1}$, $\hat{S}_{t+3}$, and $\hat{S}_{t+9}$ denote the predicted segmentation of short-, mid-, and long-term settings, corresponding to the same $X$.}
    \label{fig:three_ranges}
    \vspace{-0.2cm}
\end{figure*}

The first row shows examples where the camera is moving forward. Our model accurately captures the relative motion dynamic between the camera and all the objects in the scene.
% Move back to main submission
Our prediction results show that the right-side street-parking car segmentation moves toward right further, similar to the ground-truth. 
The second example shows that our model can capture and predict the future based on different motion patterns. In this sample, 
the camera is relatively static. However, 
the pedestrians, bike, and cars are moving toward different directions in the scene. Our model is able to distinguish each object and identify their moving directions.
% Move back to main
Specifically, a biker and a car are moving toward each other. Our model is able to predict these two segments will intersect in the future frame, but was unable to really figure out which one should be in the foreground due to the lack of depth information. This problem can be an interesting future research work, potentially solvable by including additional depth information if provided.
More qualitative results can be found in the supplementary material.

\vspace{0pt}\noindent\textbf{Short-term Forecasting:} In the short-term setting, we use the past RGB sequence $X_{15}, X_{16}, X_{17}, X_{18}$ to forecast the future semantic segmentation  $S_{19}$. Our model precisely predicts both the directions and the magnitude of the movements for the cars, as shown in Fig.~\ref{fig:short_qual}. 
% Move back to main
In the first row, the car is moving toward right, and the predicted car position in the future frame is the same as that in the ground-truth. In the second example, the left parked car is moving out of the frame due to the camera motion. Again, our model captures the exact relative motion dynamic information and precisely forecasts the same shape, size, and location of the area occupied by the same parked car in the future frame. 
Another interesting issue is that the ground-truth annotation of the second example actually misses the circle direction sign (yellow color in the segmentation map) while our model is capable to detect that direction sign segment and place it in the right position in the future frame.

\vspace{0pt}\noindent\textbf{Long-term Forecasting:} In addition, Fig.~\ref{fig:long_qual} shows the long-term forecasting results. 
Our model uses the past RGB image sequence $X_{1}, X_{4}, X_{7}, X_{10}$ as the inputs, and generates the future semantic segmentation $S_{19}$, which is 9 steps in the future from the last input frame. 
Such setting is more challenging than mid-term and short-term forecasting. However, our model can still accurately predict the moving directions of the parked cars (specially in the first example) and the pedestrians in the second example.
%In comparison with the ground-truth, 
Although our results correctly predict the moving directions, the magnitude of the movements seem to be smaller than the ground-truth. This may be due to changes in the speeds of the objects in the frames that are not observed by our model.

\vspace{0pt}\noindent\textbf{Time-horizon Comparison:} Fig.~\ref{fig:three_ranges} shows the forecasting results for our three time horizons. For all these settings, we use the same frame, namely $X_{16}$, as the last frame of the input sequence. We forecast the future segmentation at three different time horizon, namely $\hat{S}_{17}$, $\hat{S}_{19}$, and $\hat{S}_{25}$. The short-term forecasting result provides the best visual quality. From the mid-term result, we can still see that the segmentation boundary of the pedestrians are reasonable. But for the long-term result, their shapes start to deform away from regular pedestrian appearance. However, we can still see that different groups of the pedestrians are moving toward their destination in the correct directions, \eg the left most pedestrian is moving toward left, and the right most pedestrian is moving toward right.
% with reasonable amount of moving distances from our three prediction results.

\section{Conclusion}
In this paper, we proposed a single-stage end-to-end trainable model for the challenging problem of predicting future frame semantic segmentation having only observed the preceding frames RGB data. This is a practical setting for autonomous systems to directly reason about the near future based on current video data without the need to acquire any other forms of meta-data. Our proposed model for solving this task included several encoding pathways to encode the past, a temporal 3D convolution structure for capturing the scene dynamics and predictive feature learning, and finally a decoder to gradually reconstruct the future semantic segmentation. We further proposed a teacher network coupled with a distillation loss for training the network to improve the overall forecasting performance. The results on the popular Cityscapes and Apolloscape datasets indicate that our method can predict future segmentation and outperform several baseline and state-of-the-art methods.

\noindent\textbf{Acknowledgements}. This work was partially funded by Panasonic and Oppo. The authors would like to thank Jingwei Ji and Damian Mrowca for their feedback on the paper.

{\small
\bibliographystyle{ieee_fullname}
\bibliography{refs.bib}
}

%-------------------------------------------------------------------------

\clearpage

% \title{Segmenting the Future\\(Supplementary Material)}
% \maketitle
% %\maketitlesupplementary
\appendix{
\section{Supplementary Material}

 \begin{figure}[h]
     \centering
     \includegraphics[width=\linewidth]{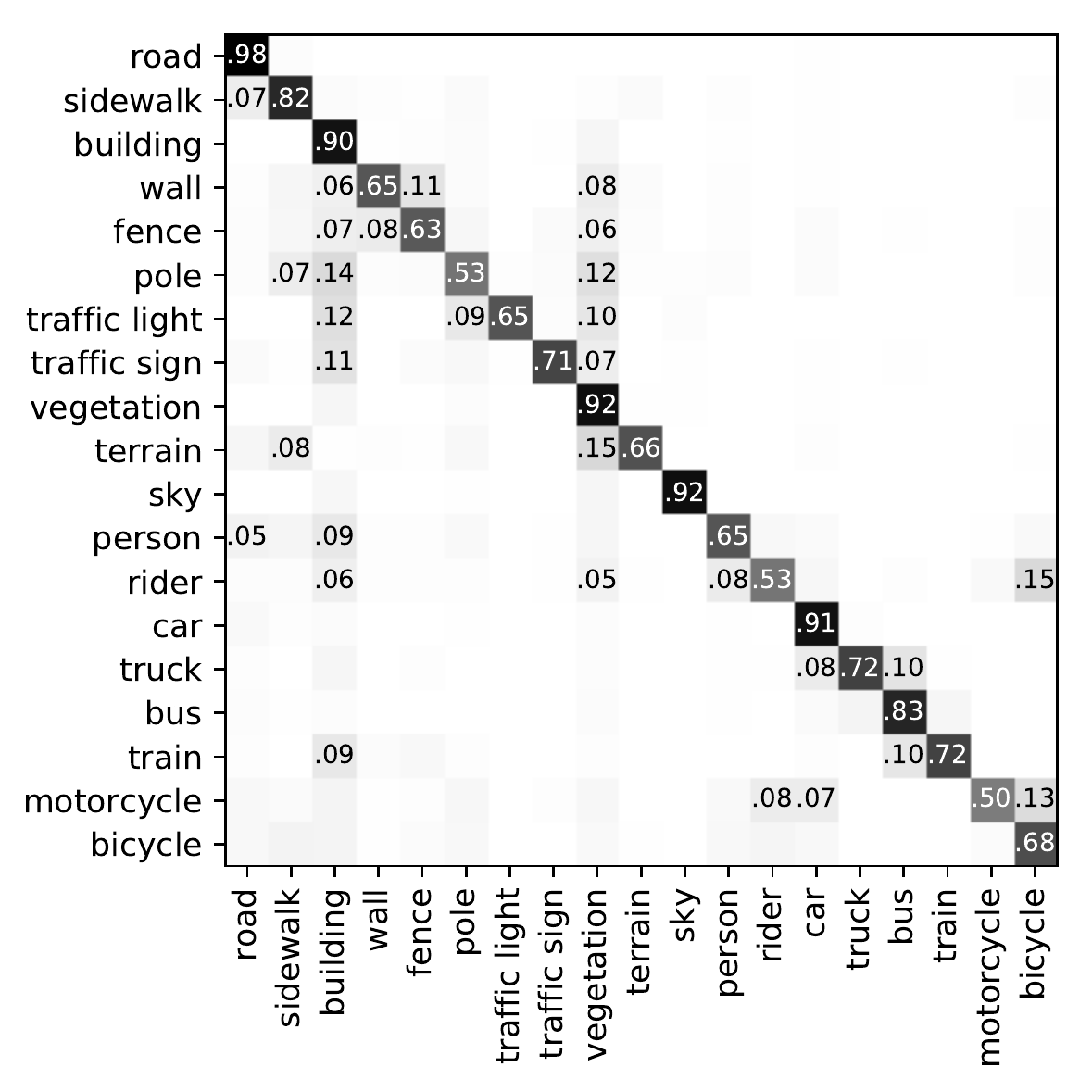}
     \caption{Confusion matrix of the mid-term semantic segmentation forecasting for all 19 classes in the Cityscapes dataset. The x-axis refers to the predicted class labels and the y-axis represents the ground-truth class labels. For instance, the confusion matrix shows that for some cases, the label `motorcycle' is misclassified as `bicycle', `rider', and `car'. %Another interesting finding is that although `building' and `vegetation' have high accuracies, several other objects belong to other classes can sometimes be misclassified to `building' and `vegetation.'
     }
     \label{fig:conf_mat}
 \end{figure}

\subsection{More Qualitative Results}
Figs.~\ref{fig:qual_short}, ~\ref{fig:qual_mid}, ~\ref{fig:qual_long} show more qualitative results for short-, mid-, and long-term forecasting, respectively. Each column represents one sample sequence, and the left-most two columns are the same samples from Fig.~\ref{fig:qual}. Here, we also show the all input frames. From the inputs, we can easily observe the movements of different objects, such as pedestrians, cars, and bikes. We can also see that our predicted future semantic segmentation follows the observed motion patterns, and our results are closer to the ground-truth future semantic segmentation for all the three time-range settings. On the contrary, predicted segmentation boundaries by ConvLSTM \cite{rochan2018future} %\cite{rochan2018future}'s 
 are not as smooth as ours. The two-stage method only works well for some examples in the short-term forecasting.%, and has difficulties in forecasting some mid- and long-term results.

\subsection{Detailed Network Architecture}
Fig.~\ref{fig:arch1} shows the detailed architecture of our proposed network.  The encoder module sequence can be seen in Table \ref{tab:arch1}. The forecasting module is a temporal 3D convolution, and the decoder structure is symmetric to the encoder.
\begin{table}[h]
  \caption{Summary of one encoder pathway based on VGG19 backbone, applied to an input frame of size $H \times W$. Feature maps are defined as the output of the last convolution module in each block. Column `\#' denotes the number of repetitions of that module. %(there is one in original VGG19, but we do not use that here.) 
  }
  \label{tab:arch1}
  \begin{center}
  {\small
  \begin{tabular}{lccc}
    \hline
    Module Sequence & \# & Resolution & Channels\\
    \hline
    Input Frame & & $H\times W$ & 3 \\
    $3\times 3$ Conv2D + BN + ReLU & 2 & $H\times W$ & 64 \\
    \hline
    Max-Pool & & $\sfrac{H}{2} \times \sfrac{W}{2}$ & 64 \\
    $3\times 3$ Conv2D + BN + ReLU & 2 & $\sfrac{H}{2} \times \sfrac{W}{2}$ & 128 \\
    \hline
    Max-Pool & & $\sfrac{H}{4} \times \sfrac{W}{4}$ & 128 \\
    $3\times 3$ Conv2D + BN + ReLU & 4 & $\sfrac{H}{4} \times \sfrac{W}{4}$ & 256 \\
    \hline
    Max-Pool & & $\sfrac{H}{8} \times \sfrac{W}{8}$ & 256 \\
    $3\times 3$ Conv2D + BN + ReLU & 4 & $\sfrac{H}{8} \times \sfrac{W}{8}$ & 512 \\
    \hline
    Max-Pool & & $\sfrac{H}{16} \times \sfrac{W}{16}$ & 512 \\
    $3\times 3$ Conv2D + BN + ReLU & 4 &$\sfrac{H}{16} \times \sfrac{W}{16}$ & 512 \\
    \hline 
  \end{tabular}
    }
  \end{center}
\end{table}

\begin{figure*}[t]
    \centering
    %\begin{tikzpicture}
    %\node[inner sep=0pt] (main) at (0,0)
    %{
    \includegraphics[width=\linewidth]{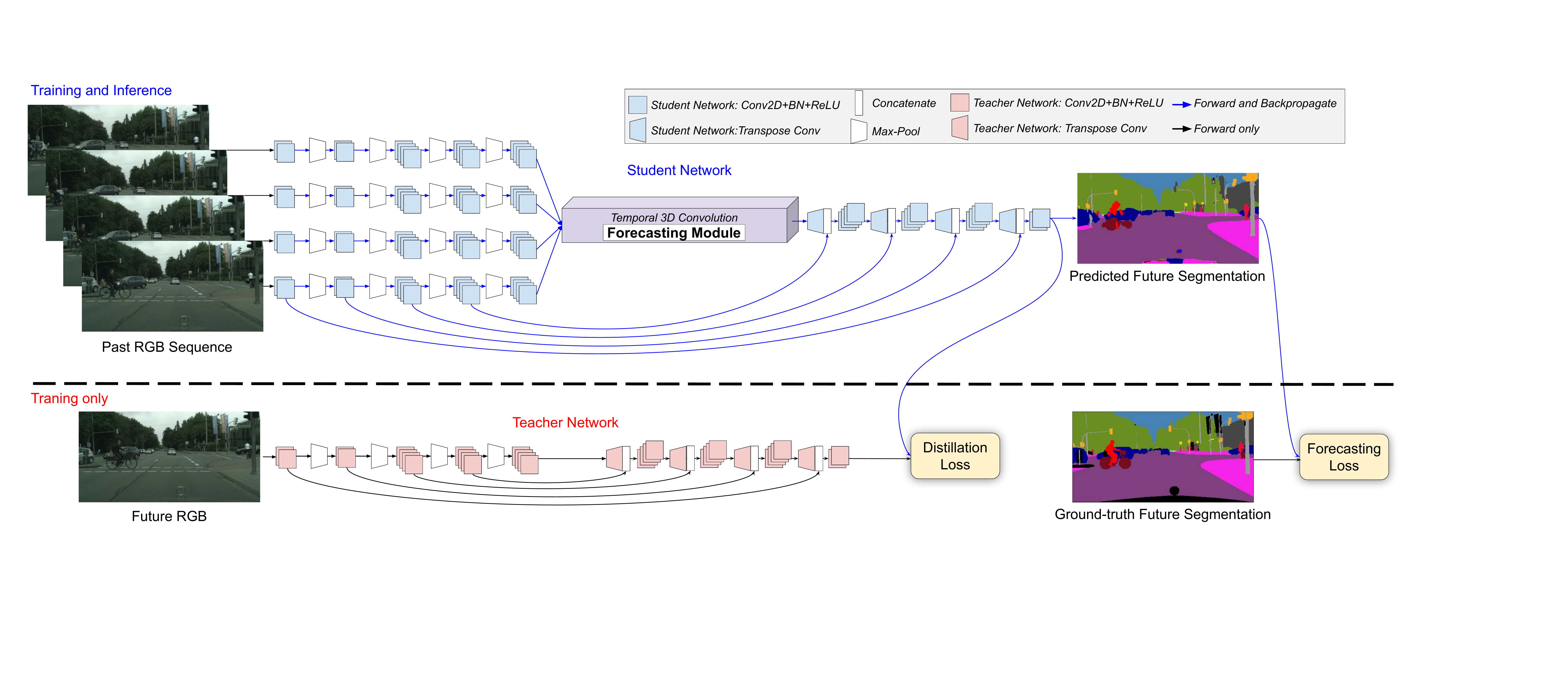}
    %};
    %\draw[] (-3.6,3.2) node[] {$X_{t-3d}$};
    %\draw[] (-2.5,3.2) node[] {$X_{t-2d}$};
    %\draw[] (-1.5,3.2) node[] {$X_{t-d}$};
    %\draw[] (-0.5,3.2) node[] {$X_{t}$};
    %\draw[] ( 3.5,3.2) node[] {$\hat{S}_{t+d'}$};
    %\end{tikzpicture}
    \caption{Network architecture overview: our student network contains encoder, forecasting module, and decoder. The encoder takes the past frames and learns latent low-dimensional representations by gradually decreasing the resolution. Our forecasting module then merges them by a temporal 3D convolution that maps them to the future. Finally, the decoder gradually increases the resolution of the output segmentation while having skip connections with the encoder at each resolution. The stacked instances of blue blocks are convolutional modules connected in a sequence. The teacher network has the same encoder and decoder architecture as the student network, but without sharing trainable parameters.
    %, similar to VGG19 \cite{simonyan2014very}.
    }
    \label{fig:arch1}
\end{figure*}

\begin{figure*}[t]
\vspace{-25pt}
\centering
%\subfigure[Short-term forecasting results]
%{
    \setlength{\tabcolsep}{2pt}
    \begin{tabular}{cccccc}
        $X_{t-3}$ &
        \raisebox{-.5\height}{\includegraphics[width=0.16\linewidth]{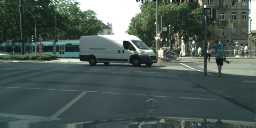}} &
        \raisebox{-.5\height}{\includegraphics[width=0.16\linewidth]{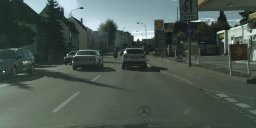}} &
        \raisebox{-.5\height}{\includegraphics[width=0.16\linewidth]{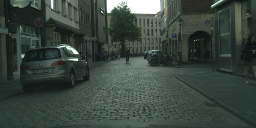}} &
        \raisebox{-.5\height}{\includegraphics[width=0.16\linewidth]{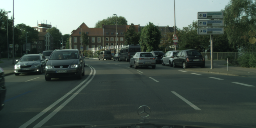}} &
        \raisebox{-.5\height}{\includegraphics[width=0.16\linewidth]{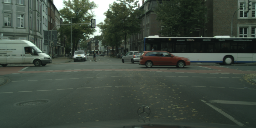}}\\
        $X_{t-2}$ &
        \raisebox{-.5\height}{\includegraphics[width=0.16\linewidth]{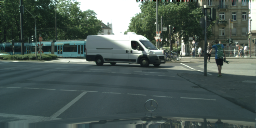}} &
        \raisebox{-.5\height}{\includegraphics[width=0.16\linewidth]{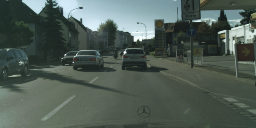}} &
        \raisebox{-.5\height}{\includegraphics[width=0.16\linewidth]{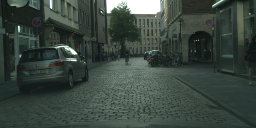}} &
        \raisebox{-.5\height}{\includegraphics[width=0.16\linewidth]{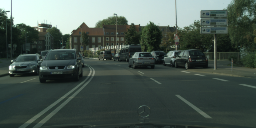}} &
        \raisebox{-.5\height}{\includegraphics[width=0.16\linewidth]{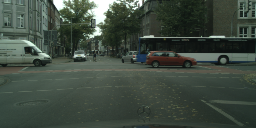}} \\
        $X_{t-1}$ &
        \raisebox{-.5\height}{\includegraphics[width=0.16\linewidth]{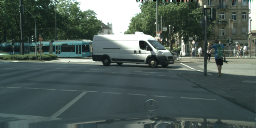}} &
        \raisebox{-.5\height}{\includegraphics[width=0.16\linewidth]{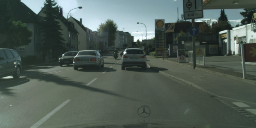}} &
        \raisebox{-.5\height}{\includegraphics[width=0.16\linewidth]{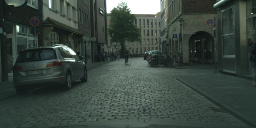}} &
        \raisebox{-.5\height}{\includegraphics[width=0.16\linewidth]{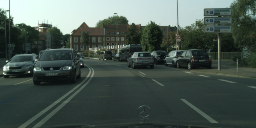}} &
        \raisebox{-.5\height}{\includegraphics[width=0.16\linewidth]{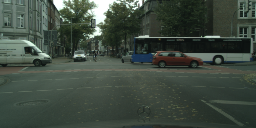}} \\
        $X_{t}$ &
        \raisebox{-.5\height}{\includegraphics[width=0.16\linewidth]{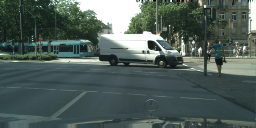}} &
        \raisebox{-.5\height}{\includegraphics[width=0.16\linewidth]{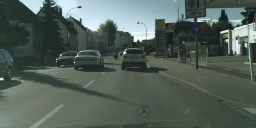}} &
        \raisebox{-.5\height}{\includegraphics[width=0.16\linewidth]{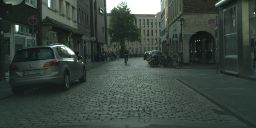}} &
        \raisebox{-.5\height}{\includegraphics[width=0.16\linewidth]{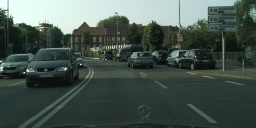}} &
        \raisebox{-.5\height}{\includegraphics[width=0.16\linewidth]{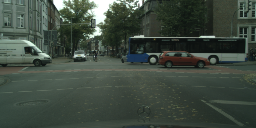}} \\
        %\rotatebox[origin=c]{90}{\scriptsize ground-truth} &
        ${S}_{t+1}$ Ground-truth &
        %\begin{tabular}{@{}c@{}}${S}_{t+3}$ \\{\scriptsize Ground-truth}\end{tabular} &
        \raisebox{-.5\height}{\includegraphics[width=0.16\linewidth]{images/short_frankurt_000001_025921/frankfurt_000001_025921_leftImg8bit_gt.png}} &
        \raisebox{-.5\height}{\includegraphics[width=0.16\linewidth]{images/short_lindau_000057_000019/lindau_000057_000019_leftImg8bit_gt.png}} &
        \raisebox{-.5\height}{\includegraphics[width=0.16\linewidth]{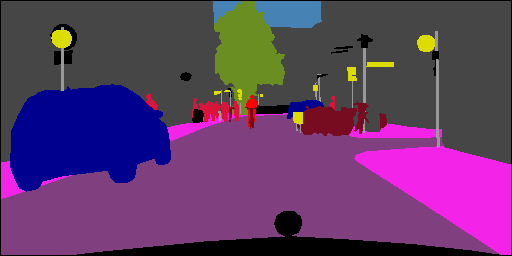}} &
        \raisebox{-.5\height}{\includegraphics[width=0.16\linewidth]{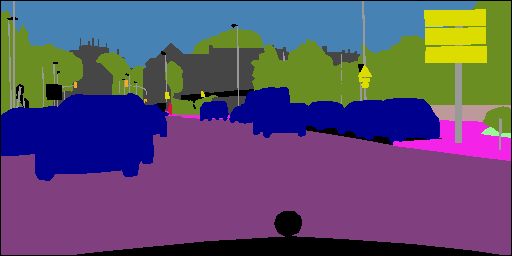}} &
        \raisebox{-.5\height}{\includegraphics[width=0.16\linewidth]{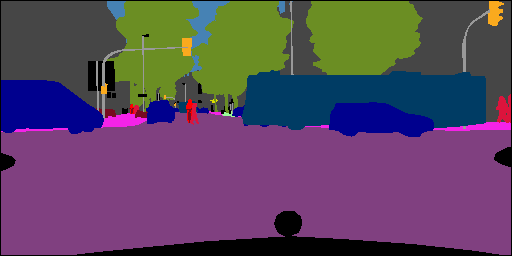}} \\
        Ours &
        \raisebox{-.5\height}{\includegraphics[width=0.16\linewidth]{images/short_frankurt_000001_025921/frankfurt_000001_025921_leftImg8bit_prediction_short.png}} &
        \raisebox{-.5\height}{\includegraphics[width=0.16\linewidth]{images/short_lindau_000057_000019/lindau_000057_000019_leftImg8bit_prediction_short.png}} &
        \raisebox{-.5\height}{\includegraphics[width=0.16\linewidth]{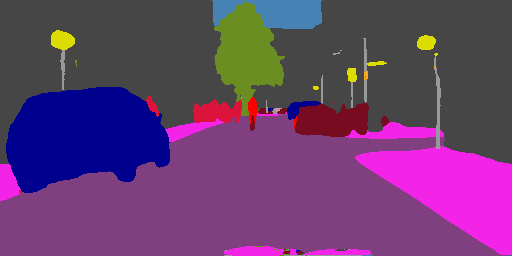}} &
        \raisebox{-.5\height}{\includegraphics[width=0.16\linewidth]{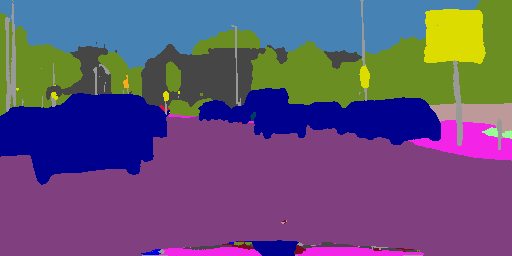}} &
        \raisebox{-.5\height}{\includegraphics[width=0.16\linewidth]{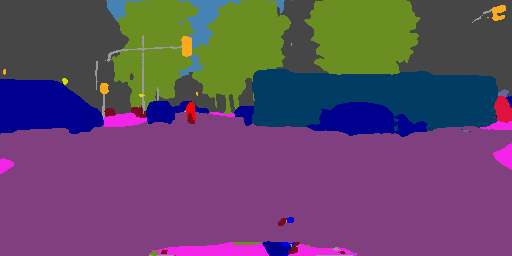}} \\
        ConvLSTM 
        &
        \raisebox{-.5\height}{\includegraphics[width=0.16\linewidth]{images/short_frankurt_000001_025921/frankfurt_000001_025921_leftImg8bit_prediction_short_conv_lstm.png}} &
        \raisebox{-.5\height}{\includegraphics[width=0.16\linewidth]{images/short_lindau_000057_000019/lindau_000057_000019_leftImg8bit_prediction_short_conv_lstm.png}} &
        \raisebox{-.5\height}{\includegraphics[width=0.16\linewidth]{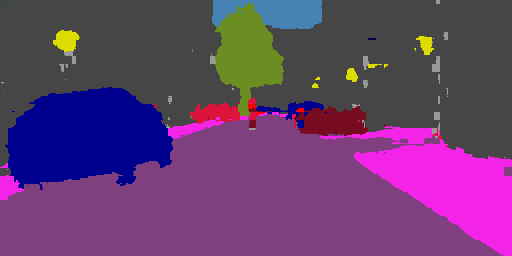}} &
        \raisebox{-.5\height}{\includegraphics[width=0.16\linewidth]{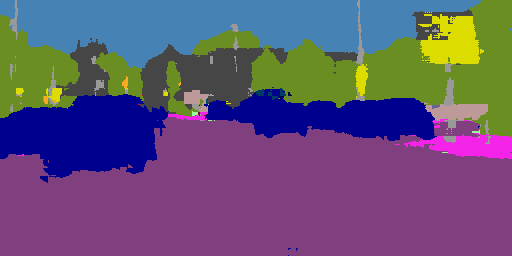}} &
        \raisebox{-.5\height}{\includegraphics[width=0.16\linewidth]{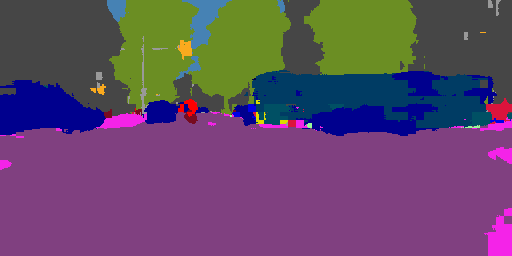}} \\
        Two-stage &
        \raisebox{-.5\height}{\includegraphics[width=0.16\linewidth]{images/short_frankurt_000001_025921/frankfurt_000001_025921_leftImg8bit_prediction_predicted_future_rgb_to_seg.png}} &
        \raisebox{-.5\height}{\includegraphics[width=0.16\linewidth]{images/short_lindau_000057_000019/lindau_000057_000019_leftImg8bit_prediction_predicted_future_rgb_to_seg.png}} &
        \raisebox{-.5\height}{\includegraphics[width=0.16\linewidth]{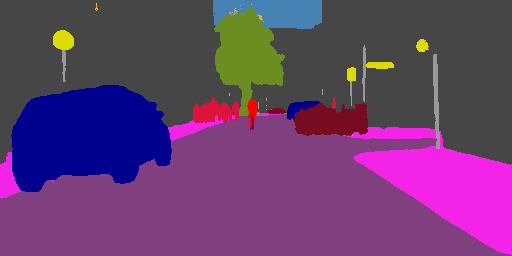}} &
        \raisebox{-.5\height}{\includegraphics[width=0.16\linewidth]{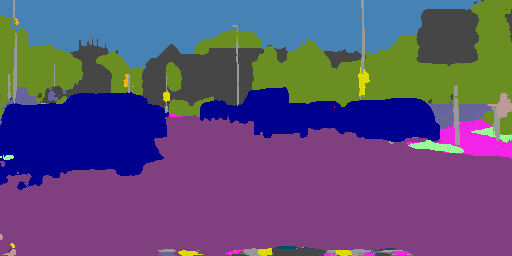}} &
        \raisebox{-.5\height}{\includegraphics[width=0.16\linewidth]{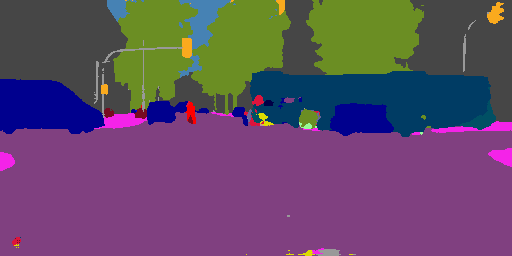}} \\
        
    \end{tabular}  
    %\label{fig:short_qual_old}
%}
    %\vspace{-5pt}
    \caption{More short-term forecasting qualitative results.}
    \label{fig:qual_short}
\end{figure*}

\begin{figure*}[t] \vspace{-25pt}
\centering
%\subfigure[Mid-term forecasting results]
%{
    \setlength{\tabcolsep}{2pt}
    \begin{tabular}{cccccc}
        $X_{t-9}$ &
        \raisebox{-.5\height}{\includegraphics[width=0.16\linewidth]{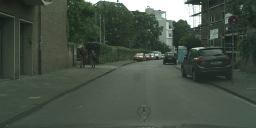}} &
        \raisebox{-.5\height}{\includegraphics[width=0.16\linewidth]{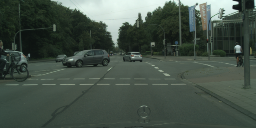}} &
        \raisebox{-.5\height}{\includegraphics[width=0.16\linewidth]{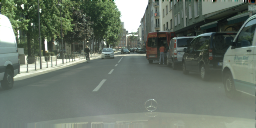}} &
        \raisebox{-.5\height}{\includegraphics[width=0.16\linewidth]{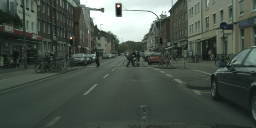}} &
        \raisebox{-.5\height}{\includegraphics[width=0.16\linewidth]{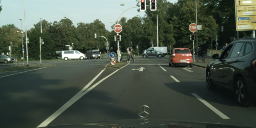}}\\
        $X_{t-6}$ &
        \raisebox{-.5\height}{\includegraphics[width=0.16\linewidth]{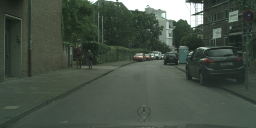}} &
        \raisebox{-.5\height}{\includegraphics[width=0.16\linewidth]{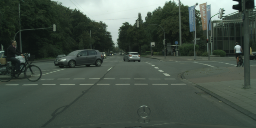}} &
        \raisebox{-.5\height}{\includegraphics[width=0.16\linewidth]{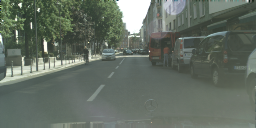}} &
        \raisebox{-.5\height}{\includegraphics[width=0.16\linewidth]{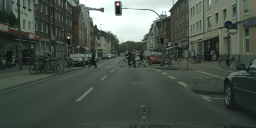}} &
        \raisebox{-.5\height}{\includegraphics[width=0.16\linewidth]{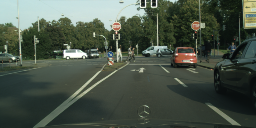}} \\
        $X_{t-3}$ &
        \raisebox{-.5\height}{\includegraphics[width=0.16\linewidth]{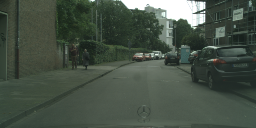}} &
        \raisebox{-.5\height}{\includegraphics[width=0.16\linewidth]{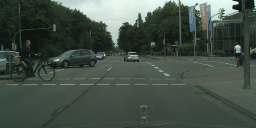}} &
        \raisebox{-.5\height}{\includegraphics[width=0.16\linewidth]{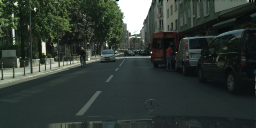}} &
        \raisebox{-.5\height}{\includegraphics[width=0.16\linewidth]{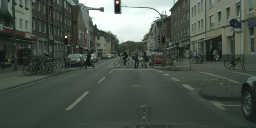}} &
        \raisebox{-.5\height}{\includegraphics[width=0.16\linewidth]{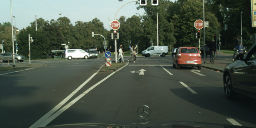}} \\
        $X_{t}$ &
        \raisebox{-.5\height}{\includegraphics[width=0.16\linewidth]{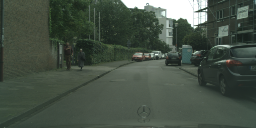}} &
        \raisebox{-.5\height}{\includegraphics[width=0.16\linewidth]{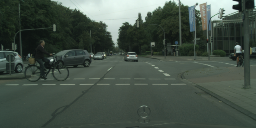}} &
        \raisebox{-.5\height}{\includegraphics[width=0.16\linewidth]{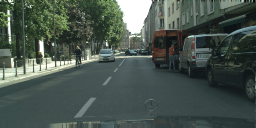}} &
        \raisebox{-.5\height}{\includegraphics[width=0.16\linewidth]{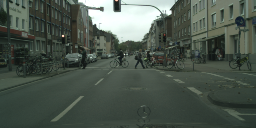}} &
        \raisebox{-.5\height}{\includegraphics[width=0.16\linewidth]{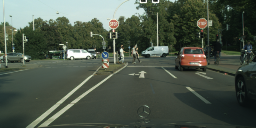}} \\
        %\rotatebox[origin=c]{90}{\scriptsize ground-truth} &
        ${S}_{t+3}$ Ground-truth &
        %\begin{tabular}{@{}c@{}}${S}_{t+3}$ \\{\scriptsize Ground-truth}\end{tabular} &
        \raisebox{-.5\height}{\includegraphics[width=0.16\linewidth]{images/mid_munster_000035_000019/munster_000035_000019_leftImg8bit_gt.png}} &
        \raisebox{-.5\height}{\includegraphics[width=0.16\linewidth]{images/mid_munster_000000_000019/munster_000000_000019_leftImg8bit_gt.png}} &
        \raisebox{-.5\height}{\includegraphics[width=0.16\linewidth]{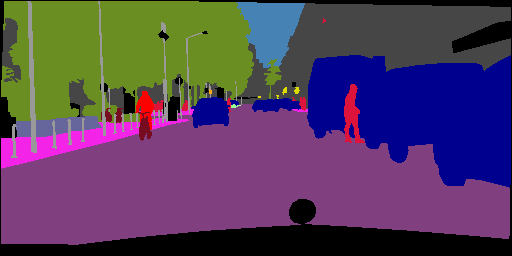}} &
        \raisebox{-.5\height}{\includegraphics[width=0.16\linewidth]{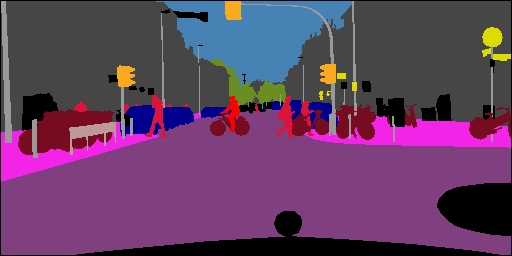}} &
        \raisebox{-.5\height}{\includegraphics[width=0.16\linewidth]{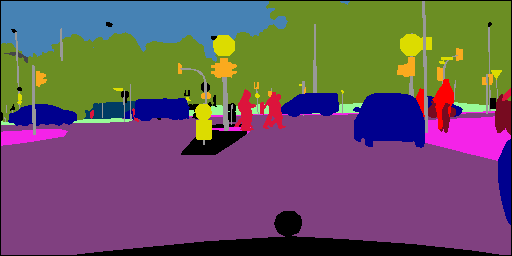}} \\
        Ours &
        \raisebox{-.5\height}{\includegraphics[width=0.16\linewidth]{images/mid_munster_000035_000019/munster_000035_000019_leftImg8bit_prediction_mid.png}} &
        \raisebox{-.5\height}{\includegraphics[width=0.16\linewidth]{images/mid_munster_000000_000019/munster_000000_000019_leftImg8bit_prediction_mid.png}} &
        \raisebox{-.5\height}{\includegraphics[width=0.16\linewidth]{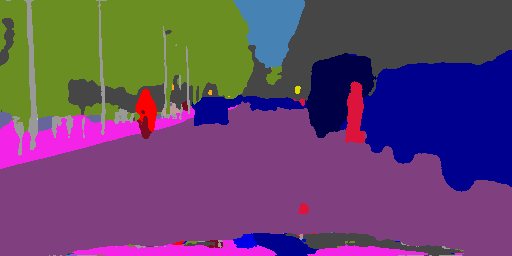}} &
        \raisebox{-.5\height}{\includegraphics[width=0.16\linewidth]{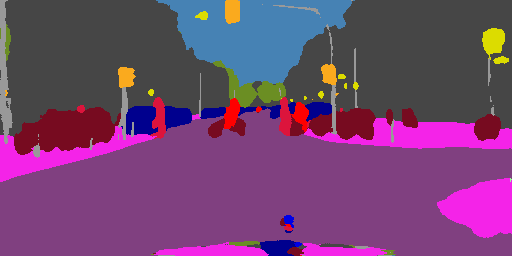}} &
        \raisebox{-.5\height}{\includegraphics[width=0.16\linewidth]{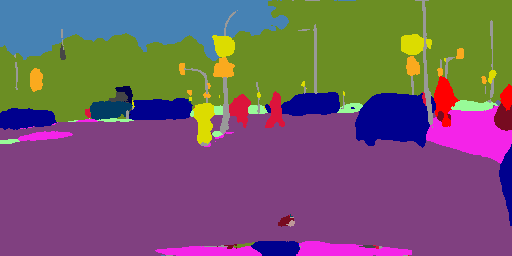}} \\
        ConvLSTM 
        &
        \raisebox{-.5\height}{\includegraphics[width=0.16\linewidth]{images/mid_munster_000035_000019/munster_000035_000019_leftImg8bit_prediction_mid_conv_lstm.png}} &
        \raisebox{-.5\height}{\includegraphics[width=0.16\linewidth]{images/mid_munster_000000_000019/munster_000000_000019_leftImg8bit_prediction_mid_conv_lstm.png}} &
        \raisebox{-.5\height}{\includegraphics[width=0.16\linewidth]{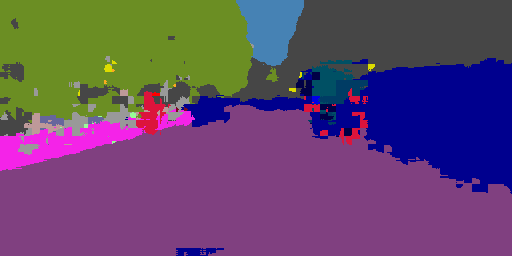}} &
        \raisebox{-.5\height}{\includegraphics[width=0.16\linewidth]{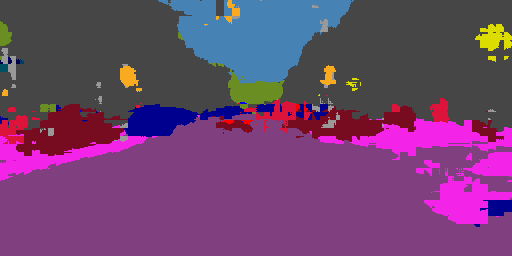}} &
        \raisebox{-.5\height}{\includegraphics[width=0.16\linewidth]{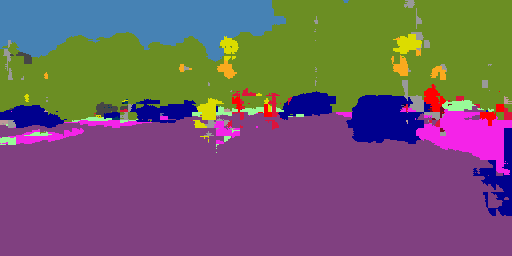}} \\
        Two-stage &
        \raisebox{-.5\height}{\includegraphics[width=0.16\linewidth]{images/mid_munster_000035_000019/munster_000035_000019_leftImg8bit_prediction_predicted_future_rgb_to_seg.png}} &
        \raisebox{-.5\height}{\includegraphics[width=0.16\linewidth]{images/mid_munster_000000_000019/munster_000000_000019_leftImg8bit_prediction_predicted_future_rgb_to_seg.png}} &
        \raisebox{-.5\height}{\includegraphics[width=0.16\linewidth]{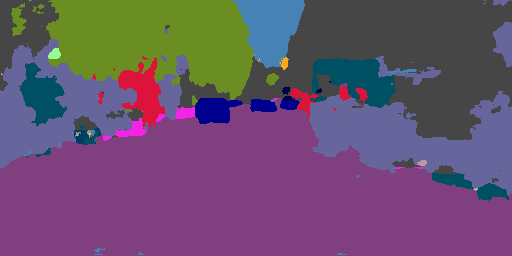}} &
        \raisebox{-.5\height}{\includegraphics[width=0.16\linewidth]{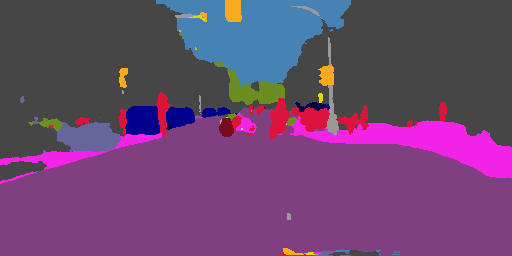}} &
        \raisebox{-.5\height}{\includegraphics[width=0.16\linewidth]{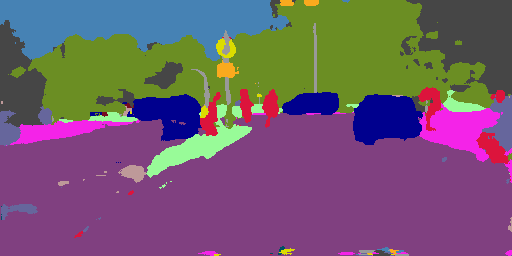}} \\
        
    \end{tabular}  
    %\label{fig:mid_qual_old}
%}
    %\vspace{-5pt}
    \caption{More mid-term forecasting qualitative results.}
    \label{fig:qual_mid}
\end{figure*}

\begin{figure*}[h!]
\centering
%\subfigure[Long-term forecasting results]
%{
    \setlength{\tabcolsep}{2pt}
    \begin{tabular}{cccccc}
        $X_{t-9}$ &
        \raisebox{-.5\height}{\includegraphics[width=0.16\linewidth]{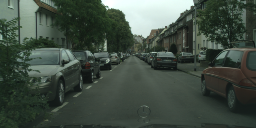}} &
        \raisebox{-.5\height}{\includegraphics[width=0.16\linewidth]{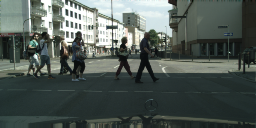}} &
        \raisebox{-.5\height}{\includegraphics[width=0.16\linewidth]{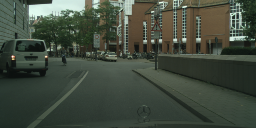}} &
        \raisebox{-.5\height}{\includegraphics[width=0.16\linewidth]{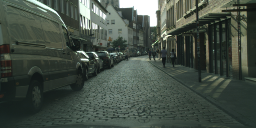}} &
        \raisebox{-.5\height}{\includegraphics[width=0.16\linewidth]{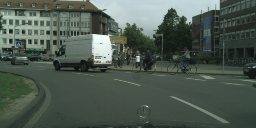}}\\
        $X_{t-6}$ &
        \raisebox{-.5\height}{\includegraphics[width=0.16\linewidth]{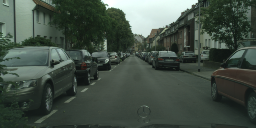}} &
        \raisebox{-.5\height}{\includegraphics[width=0.16\linewidth]{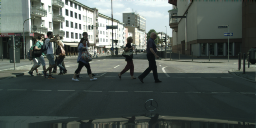}} &
        \raisebox{-.5\height}{\includegraphics[width=0.16\linewidth]{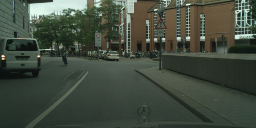}} &
        \raisebox{-.5\height}{\includegraphics[width=0.16\linewidth]{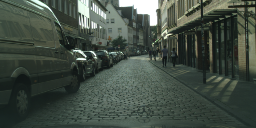}} &
        \raisebox{-.5\height}{\includegraphics[width=0.16\linewidth]{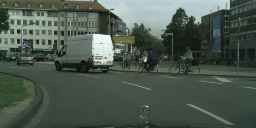}} \\
        $X_{t-3}$ &
        \raisebox{-.5\height}{\includegraphics[width=0.16\linewidth]{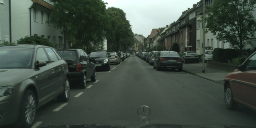}} &
        \raisebox{-.5\height}{\includegraphics[width=0.16\linewidth]{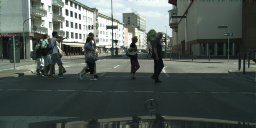}} &
        \raisebox{-.5\height}{\includegraphics[width=0.16\linewidth]{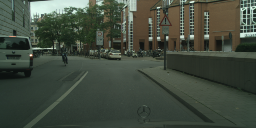}} &
        \raisebox{-.5\height}{\includegraphics[width=0.16\linewidth]{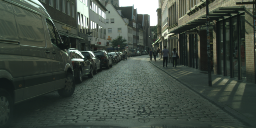}} &
        \raisebox{-.5\height}{\includegraphics[width=0.16\linewidth]{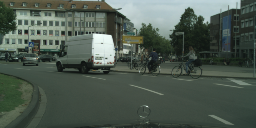}} \\
        $X_{t}$ &
        \raisebox{-.5\height}{\includegraphics[width=0.16\linewidth]{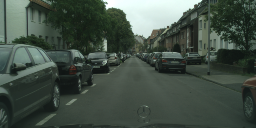}} &
        \raisebox{-.5\height}{\includegraphics[width=0.16\linewidth]{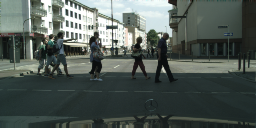}} &
        \raisebox{-.5\height}{\includegraphics[width=0.16\linewidth]{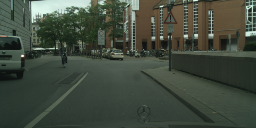}} &
        \raisebox{-.5\height}{\includegraphics[width=0.16\linewidth]{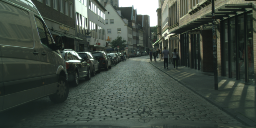}} &
        \raisebox{-.5\height}{\includegraphics[width=0.16\linewidth]{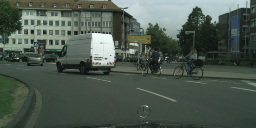}} \\
        %\rotatebox[origin=c]{90}{\scriptsize ground-truth} &
        ${S}_{t+9}$ Ground-truth &
        %\begin{tabular}{@{}c@{}}${S}_{t+3}$ \\{\scriptsize Ground-truth}\end{tabular} &
        \raisebox{-.5\height}{\includegraphics[width=0.16\linewidth]{images/long_munster_000008_000019/munster_000008_000019_leftImg8bit_gt.png}} &
        \raisebox{-.5\height}{\includegraphics[width=0.16\linewidth]{images/long_frankurt_000001_059119/frankfurt_000001_059119_leftImg8bit_gt.png}} &
        \raisebox{-.5\height}{\includegraphics[width=0.16\linewidth]{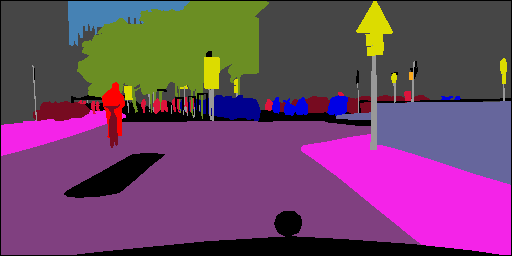}} &
        \raisebox{-.5\height}{\includegraphics[width=0.16\linewidth]{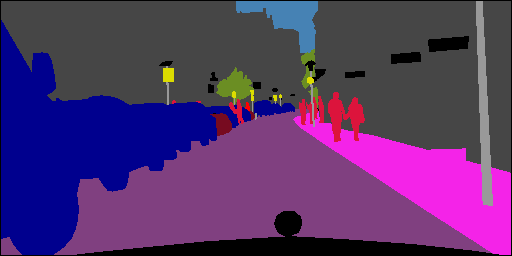}} &
        \raisebox{-.5\height}{\includegraphics[width=0.16\linewidth]{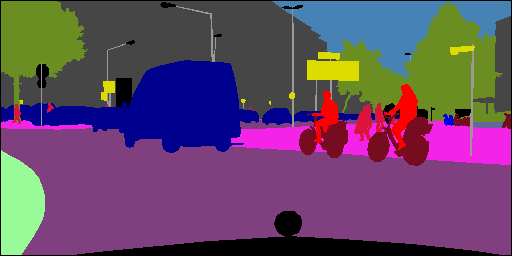}} \\
        Ours &
        \raisebox{-.5\height}{\includegraphics[width=0.16\linewidth]{images/long_munster_000008_000019/munster_000008_000019_leftImg8bit_prediction_long.png}} &
        \raisebox{-.5\height}{\includegraphics[width=0.16\linewidth]{images/long_frankurt_000001_059119/frankfurt_000001_059119_leftImg8bit_prediction_long.png}} &
        \raisebox{-.5\height}{\includegraphics[width=0.16\linewidth]{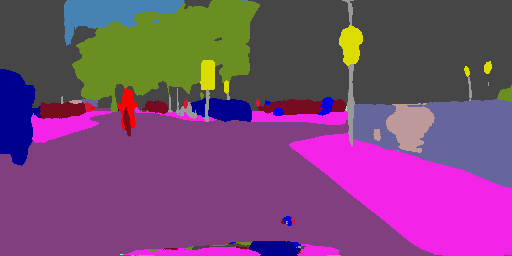}} &
        \raisebox{-.5\height}{\includegraphics[width=0.16\linewidth]{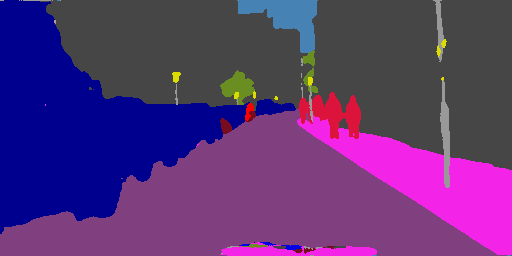}} &
        \raisebox{-.5\height}{\includegraphics[width=0.16\linewidth]{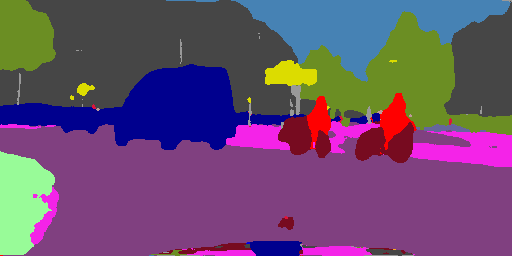}} \\
        ConvLSTM 
        &
        \raisebox{-.5\height}{\includegraphics[width=0.16\linewidth]{images/long_munster_000008_000019/munster_000008_000019_leftImg8bit_prediction_long_conv_lstm.png}} &
        \raisebox{-.5\height}{\includegraphics[width=0.16\linewidth]{images/long_frankurt_000001_059119/frankfurt_000001_059119_leftImg8bit_prediction_long_conv_lstm.png}} &
        \raisebox{-.5\height}{\includegraphics[width=0.16\linewidth]{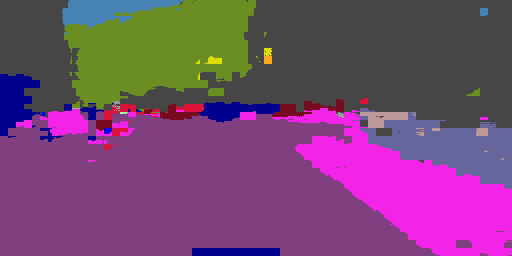}} &
        \raisebox{-.5\height}{\includegraphics[width=0.16\linewidth]{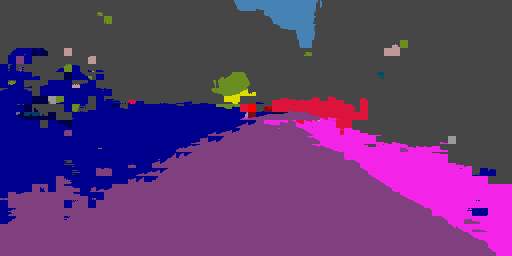}} &
        \raisebox{-.5\height}{\includegraphics[width=0.16\linewidth]{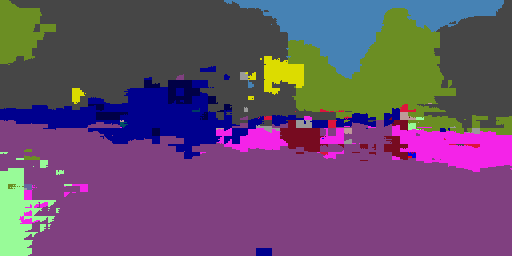}} \\
        Two-stage &
        \raisebox{-.5\height}{\includegraphics[width=0.16\linewidth]{images/long_munster_000008_000019/munster_000008_000019_leftImg8bit_prediction_predicted_future_rgb_to_seg.png}} &
        \raisebox{-.5\height}{\includegraphics[width=0.16\linewidth]{images/long_frankurt_000001_059119/frankfurt_000001_059119_leftImg8bit_prediction_predicted_future_rgb_to_seg.png}} &
        \raisebox{-.5\height}{\includegraphics[width=0.16\linewidth]{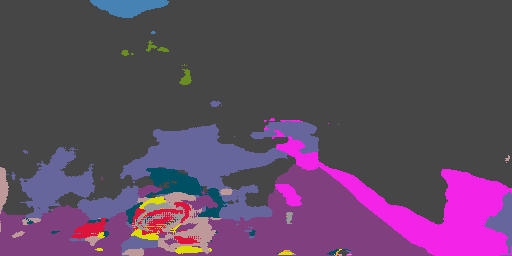}} &
        \raisebox{-.5\height}{\includegraphics[width=0.16\linewidth]{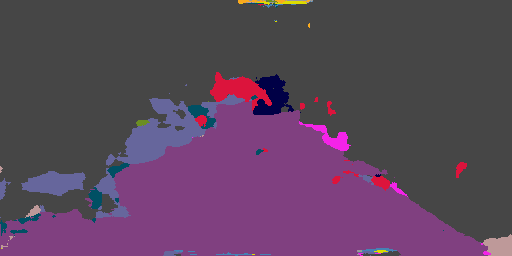}} &
        \raisebox{-.5\height}{\includegraphics[width=0.16\linewidth]{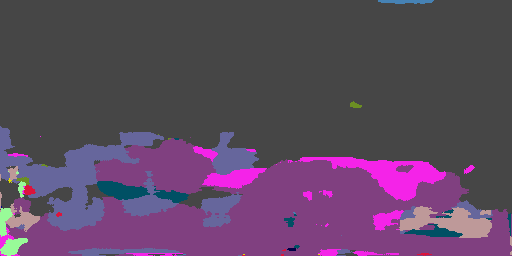}} \\
        
    \end{tabular}  
    %\label{fig:long_qual_old}
%}
    %\vspace{-5pt}
    \caption{More long-term forecasting qualitative results.}
    \label{fig:qual_long}
\end{figure*}
}

\end{document}